%% file: main.tex
\documentclass{article}

\usepackage{style/fancyhdr,style/natbib}
\usepackage{style/iclr2027_conference,times}   %

\usepackage{microtype}
\usepackage{caption}
\usepackage{subcaption}
\usepackage{xspace}
\usepackage{pifont}
\usepackage{multirow}
\usepackage{booktabs}
\usepackage{makecell}
\usepackage{diagbox}
\usepackage{longtable}
\usepackage{xltabular}
\usepackage{graphicx}
\usepackage{wrapfig}
\usepackage[table]{xcolor}   %
\usepackage{tcolorbox}
\usepackage{fontawesome5}
\usepackage{enumitem}
\usepackage{amsfonts}
\usepackage{amsmath}
\usepackage{amssymb}
\usepackage{algorithm}
\usepackage{algpseudocode}
\usepackage[bottom]{footmisc}
\usepackage{hyperref}
\usepackage{url}

\usepackage{minitoc}
\mtcsettitle{parttoc}{Contents}
\mtcsetrules{parttoc}{on}
\mtcsetfont{parttoc}{section}{\fontsize{13}{15.6}\bfseries\selectfont}
\mtcsetfont{parttoc}{subsection}{\fontsize{13}{15.6}\selectfont}
\mtcsetfeature{parttoc}{close}{\addvspace{0.8em}}

\definecolor{citecolor}{HTML}{1976D2}
\definecolor{brandgreen}{HTML}{DCEEDB}
\hypersetup{
    colorlinks=true,
    linkcolor=black,
    filecolor=magenta,
    urlcolor=blue!50!black,
    citecolor=citecolor,
}

\renewcommand{\arraystretch}{1.18}
\tcbset{
    contributionbox/.style={
        colback=blue!3,
        colframe=blue!50!black,
        boxrule=0.5pt,
        arc=3pt,
        left=6pt,
        right=6pt,
        top=6pt,
        bottom=6pt,
        fonttitle=\bfseries
    }
}

\newtcolorbox{findingbox}{
    colback=blue!3,
    colframe=blue!50!black,
    boxrule=0.5pt,
    arc=3pt,
    left=6pt,
    right=6pt,
    top=6pt,
    bottom=6pt,
}

\makeatletter
\def\@BTrule[#1]{%
  \ifx\longtable\undefined
    \let\@BTswitch\@BTnormal
  \else\ifx\hline\LT@hline
    \nobreak
    \let\@BTswitch\@BLTrule
  \else
     \let\@BTswitch\@BTnormal
  \fi\fi
  \global\@thisrulewidth=#1\relax
  \ifnum\@thisruleclass=\tw@\vskip\@aboverulesep\else
  \ifnum\@lastruleclass=\z@\vskip\@aboverulesep\else
  \ifnum\@lastruleclass=\@ne\vskip\doublerulesep\fi\fi\fi
  \@BTswitch}
\makeatother

\input{style/commands.tex}

\newcommand{\growmtp}{\textit{GrowMTP}\xspace}
\newcommand{\growmtpgrad}{%
\textcolor[HTML]{21A04C}{G}%
\textcolor[HTML]{219661}{r}%
\textcolor[HTML]{228B76}{o}%
\textcolor[HTML]{22818B}{w}%
\textcolor[HTML]{236CB6}{M}%
\textcolor[HTML]{2461CB}{T}%
\textcolor[HTML]{2457E0}{P}}

\title{\growmtpgrad: Can RL Grow Its Own Draft Head?}

\author{
Minghua He$^{1,2}$\thanks{This work was done during the internship at WeChat AI.},
Lingzhe Zhang$^{2}$,
Yuan Liu$^{1}$,
Xiao Zhou$^{1}$,
Aiwei Liu$^{1}$\thanks{Corresponding author.} \\
$^{1}$WeChat AI, Tencent,
$^{2}$Peking University \\
\texttt{hemh2120@stu.pku.edu.cn, coveliu@tencent.com}
}

\iclrfinalcopy   %

\begin{document}

\maketitle
\lhead{GrowMTP: Can RL Grow Its Own Draft Head?}   %

\doparttoc          %
\faketableofcontents %

\input{sections/abstract}

\input{sections/introduction}
\input{sections/preliminaries}
\input{sections/method}
\input{sections/experiments}
\input{sections/related_work}
\input{sections/conclusion}
\clearpage

\bibliographystyle{style/iclr2027_conference}
\bibliography{style/iclr2027_conference}

\clearpage
\appendix

\renewcommand{\thepart}{}
\renewcommand{\partname}{}
\part{Appendix}
\parttoc
\clearpage

\input{sections/appendix}

\end{document}

%% file: style/commands.tex
\renewcommand{\epsilon}{\varepsilon}

%% file: sections/abstract.tex
\begin{abstract}
Reinforcement learning (RL) post-training drives the frontier capabilities of large language models, with its wall-clock dominated by autoregressive rollout generation.
Speculative decoding is an established remedy for this bottleneck, but existing draft heads must be pretrained or warmed up before RL, introducing substantial training cost outside the RL run to be accelerated.
We observe that RL training itself provides both conditions required for online draft-head training: its rollout distribution is far narrower than that of pretraining, and its verification step continuously produces supervision signals aligned with this distribution.
Building on these observations, we propose \growmtp, which uses this supervision to train a draft head from scratch entirely within the RL loop, with all head updates detached from the policy backbone.
On Qwen3-4B (no draft head), MiMo-7B-SFT (weak head), and Qwen3.5-4B-Base (strong head), \growmtp achieves rollout speedups of $2.13\times$, $1.93\times$, and $1.36\times$, and end-to-end speedups of $1.60\times$, $1.41\times$, and $1.20\times$, respectively.
\growmtp therefore serves existing RL training frameworks as a modular component, particularly offering a from-scratch acceleration path for models without pretrained draft heads.
Website: {\hypersetup{urlcolor=citecolor}\textbf{\href{https://growmtp.github.io/}{https://growmtp.github.io/}}}.
\end{abstract}

%% file: sections/introduction.tex
\section{Introduction}
\label{sec:intro}

\begin{figure}[h]
    \centering
    \includegraphics[width=0.85\linewidth]{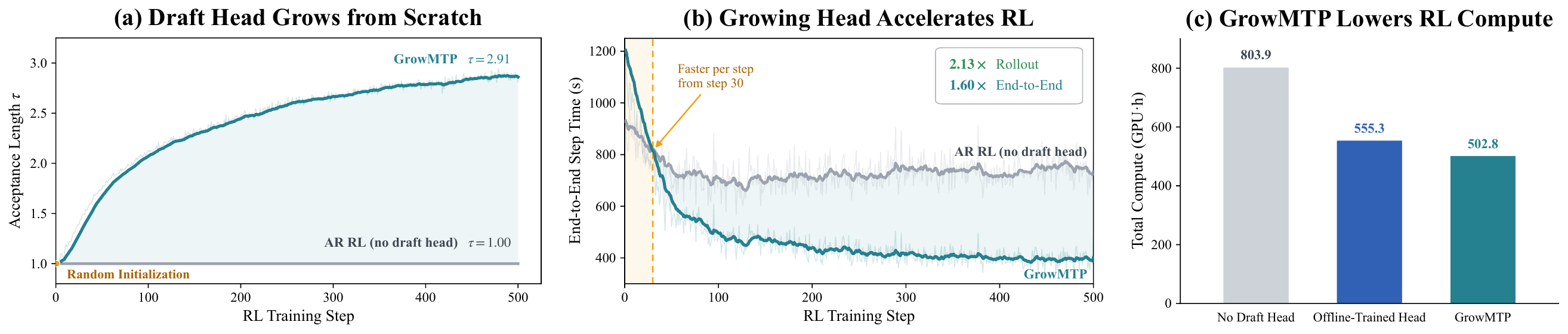}
    \vspace{-4pt}
    \caption{\textbf{An RL run can grow its own draft head from scratch and reduce its training compute.}
    \textbf{(a)} The acceptance length grows from $1.00$ to $2.91$.
    \textbf{(b)} This yields $2.13\times$ rollout and $1.60\times$ end-to-end speedups, including head updates.
    \textbf{(c)} \growmtp uses less total compute than matched offline-trained and no-head baselines.
    Qwen3-4B, $500$ RL steps on DAPO-Math-17K.}
    \label{fig:teaser}
    \vspace{-0.2cm}
\end{figure}

Reinforcement learning (RL) post-training now drives the frontier capabilities of large language models~\citep{jaech2024openai,guo2025deepseek}.
The RL stage accounts for a substantial share of the training budget~\citep{khatri2026art}: OpenAI o3 was trained with $10\times$ the RL compute of o1~\citep{singh2025openai}.
Within this budget, the dominant cost is rollout generation: the policy decodes tens of thousands of tokens per response token by token, typically 70--90\% of each training step~\citep{chen2026respec}.
The bottleneck is thus not the RL algorithm itself but autoregressive decoding.

This bottleneck is not unique to RL: it has long been central to model serving, where speculative decoding is the established solution~\citep{leviathan2023fast,chen2023accelerating}.
A cheap draft proposes several tokens, the target model verifies them in a single parallel forward pass, and rejection sampling preserves the output distribution exactly.
Modern drafts are lightweight auxiliary~\citep{ankner2024hydra,li2024eagle,li2024eagle2} or native multi-token-prediction (MTP) heads~\citep{cao2026qwen3,xiaomi2025mimo}, which reuse the backbone's hidden states at negligible cost and are now standard components of frontier models.
Porting this machinery to RL rollouts is therefore a natural step.

However, existing approaches train draft heads before RL~\citep{cai2024medusa}.
An EAGLE-style head for a 70B target model takes roughly 5,000 offline GPU hours~\citep{li2026eagle,hu2026bridging}, while native MTP heads are trained throughout pretraining, 14.8T tokens for DeepSeek-V3~\citep{liu2024deepseek}.
Improvement is no different: drafting depth depends on training, and raising acceptance at larger $K$ typically requires another offline stage.
Drafting capability thus enters RL as a prerequisite, and a model without a head has no route to acceleration.
In effect, the established remedy for the dominant cost of RL is itself a further training cost, incurred before RL benefits at all.

We argue that this tension stems from a twofold mismatch between how draft heads are trained and what an RL process both requires and provides.
Draft heads brought into RL are pretrained for broad deployment across diverse tasks.
What RL requires is much narrower: rollouts concentrate on a particular task distribution, and the head only needs to fit it.
Meanwhile, existing pipelines source draft-head supervision entirely outside RL.
Yet during speculative rollouts, the target model verifies every draft, producing predictive distributions and acceptance decisions that can supervise the draft head but are otherwise discarded.
An RL process can thus grow its own draft head from the rollouts it already requires and accelerate subsequent rollouts, without an upfront training bill.

Although RL provides the distribution and the supervision, training a multi-step draft head online presents three challenges.
First, draft-head training must remain consistent with rollout-time drafting.
Each proposal is conditioned on the preceding drafts, so its verification signal must be applied under the same draft-conditioned state.
Second, learning across drafting depths is inherently dependent.
The $k$-th draft token counts only if the preceding $k-1$ are accepted, so early rejections limit gains from deeper predictions.
Training should therefore reflect this dependence when assigning supervision across drafting depths.
Finally, not every verification signal corresponds to an executed prefix.
Once the target rejects a draft token, deeper drafts remain conditioned on it and no longer follow the executed sequence, so supervising all positions equally also trains the head on discarded suffixes.

To meet these demands, we introduce \growmtp, which trains the draft head online from rollout verification signals.
To preserve consistency with rollout-time drafting, \growmtp records the draft trajectory and reconstructs its draft-conditioned states, applying each signal under the state that produced it.
To encode the dependence across drafting depths, we design a depth-coupled acceptance (DCA) loss based on an acceptance-chain surrogate.
We apply a logarithmic transformation to give cycles with smaller chain values greater relative gradient weight.
Because this objective can still assign nonzero gradients beyond the first rejection, \growmtp gates the loss there and excludes subsequent positions.
All draft-head updates are detached from the policy backbone, without directly interfering with policy learning.
Since the first draft position always provides valid supervision, training can begin from random initialization without an offline stage or a depth schedule.

To evaluate from-scratch training, we train a random $K{=}5$ head on Qwen3-4B~\citep{yang2025qwen3} for 500 math RL steps.
Its acceptance length grows from the autoregressive floor to 2.91, accelerating rollouts by $2.13\times$ and end-to-end steps by $1.60\times$, with early per-step speedups.
Over 500 steps, \growmtp reduces total GPU-hours by $9.4\%$ versus an offline-trained head at comparable acceptance.
\growmtp also accelerates models with pretrained heads: on Qwen3.5-4B-Base~\citep{qwen35blog}, online updates accelerate rollouts by $1.36\times$ over a frozen head; on MiMo-7B-SFT~\citep{xiaomi2025mimo}, extending the head from $K{=}1$ to $K{=}5$ raises acceptance from 1.65 to 4.0 and accelerates rollouts by $1.93\times$.
Heads trained on Math and Code rollouts achieve higher in-domain acceptance.
Thus, in-RL draft-head training does not replace MTP pretraining but suffices for the current RL run.

In summary, our contributions are as follows:
\begin{itemize}
    \item \textbf{Revisiting MTP for RL acceleration.} We show that an RL run only requires a draft head adapted to its rollout distribution and continuously produces matching verification supervision, allowing such a head to be trained during RL rather than pretrained.
    \item \textbf{GrowMTP.} We propose \growmtp, which trains an MTP head online using valid verification supervision under rollout-consistent draft states, extends learning along the accepted draft chain, and retains the original RL objective.
    \item \textbf{From-scratch acceleration.} On the headless Qwen3-4B, \growmtp grows a random $K{=}5$ head for a $2.13\times$ rollout and $1.60\times$ end-to-end speedup, head-training cost included.
    \item \textbf{Modular RL acceleration.} \growmtp extends speculative acceleration to models without pretrained MTP heads, which grow one during RL instead.
\end{itemize}

%% file: sections/preliminaries.tex
\section{Preliminaries}
\label{sec:prelim}

\subsection{Reinforcement Learning for Large Language Models}
\label{sec:prelim-rl}

We consider Group Relative Policy Optimization (GRPO)-style RL post-training~\citep{gspo}, in which the policy $\pi_\theta$, the language model itself, samples a group of $G$ responses for each prompt $x$ and receives a scalar reward per response.
Each response $y$ is generated autoregressively: the policy emits one token per forward pass, $y_t \sim \pi_\theta(\cdot \mid x, y_{<t})$.
A training step thus comprises two phases: \emph{rollout}, which generates the responses, and \emph{update}, which converts the rewards into advantages and adjusts the policy.
The cost of a step is dominated by rollout, because of this token-by-token decoding, and this work accordingly targets the rollout bottleneck while retaining the original RL objective.

\subsection{Speculative Decoding and Multi-Token Prediction}
\label{sec:prelim-specdec}

Speculative decoding generates tokens in \emph{draft-then-verify} cycles. We use a recurrent MTP head~\citep{xu2026deepseek} $h_\phi$ as the draft head, reusing backbone hidden states to propose $K$ tokens per cycle; $K$ is the \emph{drafting depth}.
Drafting is itself autoregressive: the $k$-th draft token $d_k$ is sampled from a head distribution $q_k(\cdot \mid x, y_{\le t}, d_{<k})$, conditioned on the committed context and on the draft tokens before it.
The target model then verifies all $K$ candidates in a single forward pass, producing its own distribution $p_k$ at every draft position, and accepts $d_k$ by \emph{rejection sampling}, with probability
\begin{equation}
    \min\!\left(1,\ \frac{p_k(d_k)}{q_k(d_k)}\right).
    \label{eq:accept-rule}
\end{equation}
At the first rejection, a replacement token is resampled from a corrected distribution, and if all $K$ candidates are accepted, a bonus token is drawn from the target's next-position distribution $p_{K+1}$, so the committed sequence remains distributed exactly as if the target had decoded autoregressively~\citep{xia2024unlocking,miao2024specinfer}.
The tokens committed in one cycle, the accepted prefix plus this final token, always number at least one: even a draft rejected at the first position commits the resampled token, regardless of the quality of the draft.
We call the average number of tokens committed per cycle the \emph{acceptance length} $\tau \in [1, K+1]$: rollout throughput grows in proportion to $\tau$, up to the drafting overhead, with $\tau = 1$ recovering plain autoregressive decoding.

\subsection{Training Draft Heads from RL Rollouts}
\label{sec:prelim-rollout-supervision}

We observe that a draft head's ability to accelerate a given RL process depends on its acceptance on that process's rollout distribution, rather than its general drafting capability.
At the same time, every speculative rollout produces target distributions and acceptance decisions on this distribution, verification supervision for exactly the proposals generated.
The first draft position always provides supervision, 
so head training can begin from random initialization using the rollout records that RL already produces for policy optimization, without a separate dataset or an offline head-training stage.

However, using this supervision effectively for multi-step online training presents three challenges.
First, each verification signal corresponds to a proposal generated under a particular draft-conditioned state, so training must apply the signal under the same state.
Second, position $k$ increases the acceptance length only if the preceding $k-1$ drafts are accepted, so training should reflect this dependence across depths.
Finally, once the target rejects a draft token, subsequent drafts remain conditioned on it and no longer follow the executed sequence; supervising all positions equally also trains the head on discarded suffixes that were drafted but never committed to the executed response.

%% file: sections/method.tex
\section{\growmtp: Growing Draft Heads within RL}
\label{sec:method}

\begin{figure}[t]
    \centering
    \includegraphics[width=0.95\linewidth]{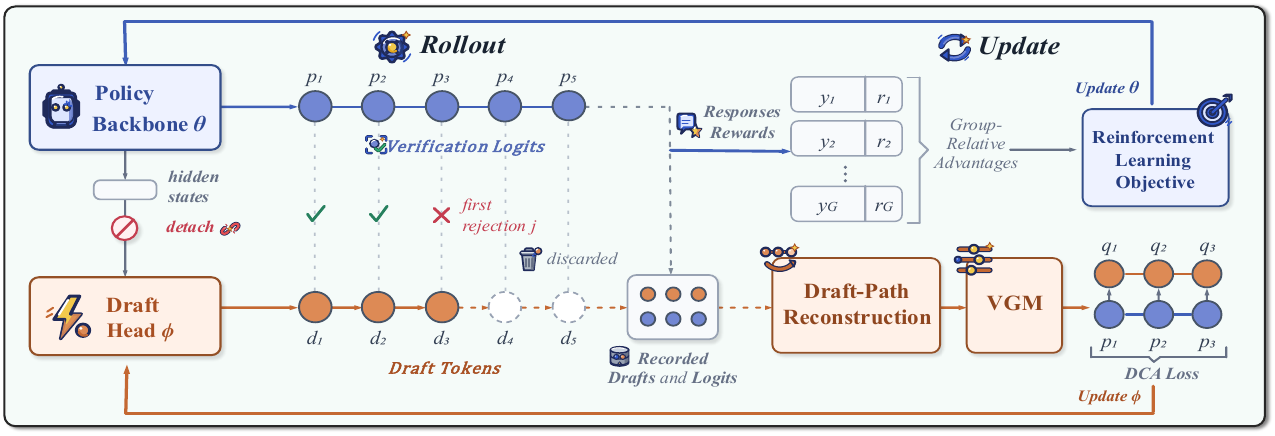}
    \caption{\textbf{The \growmtp pipeline.} Rollout performs speculative decoding as usual and records the drafts, verification logits, and acceptance decisions of each cycle. Update trains the draft head on these records through detached hidden states, alongside the policy update.}
    \label{fig:method}
    \vspace{-0.4cm}
\end{figure}

\growmtp uses verification supervision from each rollout to train the draft head for subsequent rollouts, forming a training-and-acceleration loop within RL.
Figure~\ref{fig:method} illustrates both phases: rollout records the realized draft trajectory and verification signals, while update trains the head from detached backbone states alongside the policy update.
Using these records, \growmtp reconstructs the draft-conditioned state under which each signal was produced, optimizes an acceptance-chain surrogate with the DCA loss, and excludes positions beyond the first rejection.
The head loss updates only draft-head parameters, while the backbone optimizes the original RL objective.
\textbf{\autoref{app:algo}} presents the full training algorithm, and \textbf{\autoref{app:impl}} describes its implementation.

\subsection{Training Pipeline and Draft-Path Reconstruction}
\label{sec:method-pipeline}

\noindent\textbf{Training Pipeline.}
\growmtp preserves the two phases of the existing RL training step: rollout performs speculative decoding as usual, and update continues the original policy optimization.
During each draft-then-verify cycle, rollout records the backbone hidden states at the start of the cycle, the drafts generated by the head, the target verification logits, and the acceptance decision at each position.
These quantities are already produced by drafting and verification, so collecting the training supervision requires no additional target-model forward pass.
During update, the original RL objective continues to update the policy, while these rollout records separately train the draft head.
The recorded backbone hidden states enter the head detached, so the head loss updates only the head parameters $\phi$ and sends no gradient to the backbone parameters $\theta$.
Figure~\ref{fig:method} illustrates this pipeline.

\begin{figure}[t]
    \centering
    \includegraphics[width=0.95\linewidth]{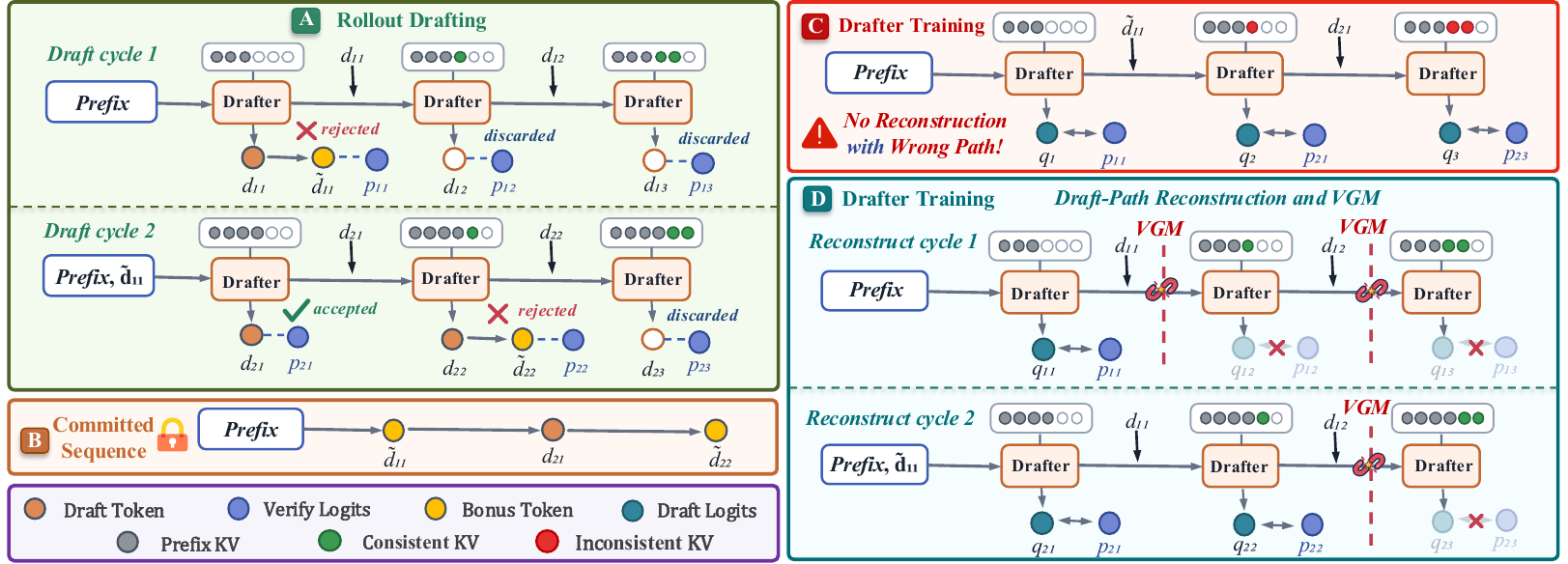}
    \caption{\textbf{Rollout cycles and the two training paths.} Teacher forcing (C) trains on the committed path. Reconstruction (D) feeds the recorded drafts back, and VGM truncates at the first rejection.}
    \label{fig:cycle}
    \vspace{-0.4cm}
\end{figure}

\noindent\textbf{Draft-Path Reconstruction.}
A natural use of these records treats the final response as an ordinary token sequence and trains the head on it by a right shift, which is teacher forcing~\citep{gloeckle2024better}, constructed in full in \textbf{\autoref{app:tf}}.
Figure~\ref{fig:cycle} contrasts the two training paths.
Consider draft-then-verify cycle $r$: its committed context $\mathbf{c}_r$ is the prompt plus all tokens committed earlier, and its draft path $\mathbf{d}_r=[d_{r,1},\ldots,d_{r,K}]$ its $K$ draft tokens. During verification, the target model $z_\theta$ conditions its logits on $\mathbf{c}_r$ and the preceding drafts, yielding the target distribution at position $k$:
\begin{equation}
    p_{r,k}=\operatorname{softmax}\bigl(z_\theta(\mathbf{c}_r,d_{r,1},\ldots,d_{r,k-1})\bigr).
    \label{eq:verification-context}
\end{equation}

Teacher forcing conditions position $k$ on the committed token rather than the head's own draft: the two are identical only when that draft was accepted.
If position $j$ is the first rejection, the target replaces $d_{r,j}$ with $\tilde{d}_{r,j}\neq d_{r,j}$, and the tokens that follow, written $\tilde{d}_{r,j+1},\ldots$ and committed by subsequent cycles, continue from this replacement, so the paths diverge as
\begin{equation}
    \mathbf{d}_r^{\mathrm{draft}}
    = [d_{r,<j},\, d_{r,j},\, d_{r,j+1},\ldots],
    \qquad
    \mathbf{d}_r^{\mathrm{commit}}
    = [d_{r,<j},\, \tilde{d}_{r,j},\, \tilde{d}_{r,j+1},\ldots].
\label{eq:draft-path-mismatch}
\end{equation}
The head consumes at each position the hidden state and token embedding of the position before it, so the committed path supplies a different pair at every position past $j$, along with different cached keys and values: teacher forcing fits the head at a state that drafting reaches only when the whole prefix is accepted.
The training state at position $k$ agrees with the drafting state if and only if none of the preceding $k-1$ positions was rejected, and this condition tightens monotonically in $k$.
The bias of teacher forcing therefore grows with draft depth, and drafting depth is what \growmtp extends during RL.
We discuss this train--inference inconsistency in detail in \textbf{\autoref{app:tf}}.

The final response neither marks cycle boundaries nor retains the drafts the target replaced, so both must be recorded during rollout.
\growmtp therefore applies \emph{Draft-Path Reconstruction}, which starts from the recorded cycle context and feeds the recorded drafts back in their original order:
\begin{equation}
    s_{r,1}=\mathbf{c}_r,
    \qquad
    s_{r,k+1}=[s_{r,k},d_{r,k}],
    \qquad
    q_{r,k}=h_\phi(\cdot\mid s_{r,k}).
    \label{eq:draft-path-reconstruction}
\end{equation}
This recursion gives $s_{r,k}=[\mathbf{c}_r,d_{r,1},\ldots,d_{r,k-1}]$, pairing every $p_{r,k}$ with a head distribution under the same cycle context and realized draft prefix.
We provide the implementation details of the reconstruction in \textbf{\autoref{app:impl}}.
The only remaining asymmetry lies between successive RL steps: the head learns from logits of the current rollout but drafts for a policy one update newer, and \textbf{\autoref{app:lag}} bounds the resulting discrepancy and shows it remains small in our runs.

\subsection{The Depth-Coupled Acceptance Loss}
\label{sec:method-dca}

To target acceleration, we construct a differentiable surrogate for the expected accepted draft length from the reconstructed pairs.
The building block is the acceptance probability at a single position: writing $p_k$ and $q_k$ for the paired target and head distributions at position $k$ within a cycle, rejection sampling accepts a token sampled from the head with probability averaged over possible draft tokens:
\begin{equation}
    \alpha_k \;=\; \sum_{d} \min\bigl(p_k(d),\, q_k(d)\bigr) \;=\; 1 - \mathrm{TV}(p_k,\, q_k),
    \label{eq:alpha}
\end{equation}
where $\mathrm{TV}$ is the total variation distance.
Acceptance is chain-dependent: the $l$-th token contributes only when positions $1$ through $l$ are all accepted.
Chaining these local rates yields a differentiable surrogate for the accepted length, whose negation is the TV loss~\citep{zhou2024distillspec,li2026breaking}
\begin{equation}
    \mathcal{L}_{\mathrm{TV}} \;=\; -\sum_{l=1}^{K} \prod_{i=1}^{l} \alpha_i.
    \label{eq:chain}
\end{equation}
To give cycles with smaller chain values greater relative gradient weight, we apply a logarithmic transformation to the acceptance-chain surrogate to obtain the DCA training objective (\textbf{\autoref{app:dca}}):
\begin{equation}
    \mathcal{L}_{\mathrm{DCA}} \;=\; -\log\bigl(-\mathcal{L}_{\mathrm{TV}}\bigr) \;=\; -\log \sum_{l=1}^{K} \exp\Bigl(\sum_{i=1}^{l} \log \alpha_i\Bigr),
    \label{eq:dca}
\end{equation}
minimized with respect to the head alone: the gradient flows through each $q_k$, and the recorded $p_k$ stay constant.
The logarithm preserves the gradient direction on each fixed cycle but changes the relative cycle weights in the batch, so the two batch objectives can have different stationary points.

\subsection{Learning up to the First Rejection}
\label{sec:method-valid}

After the first rejection, later draft positions remain conditioned on discarded tokens and no longer follow the executed sequence.
Verification still produces target distributions at each of these draft positions, so treating every position equally also assigns full training weight to the discarded suffix.

However, differentiating DCA with respect to $\alpha_k$ factors the gradient weight at position $k$ into the product of the preceding acceptance probabilities and a bounded tail term,
\begin{equation}
    -\frac{\partial \mathcal{L}_{\mathrm{DCA}}}{\partial \alpha_k} \;=\; \Bigl(\prod_{i<k} \alpha_i\Bigr) \frac{D_k}{-\mathcal{L}_{\mathrm{TV}}},
    \qquad
    D_k \;=\; \sum_{l=k}^{K} \prod_{i=k+1}^{l} \alpha_i \;\in\; [1,\, K-k+1],
    \label{eq:weight}
\end{equation}
where beyond the leading product, its denominator shared across positions, the weights differ at most by a factor of $K$, with the full derivation given in \textbf{\autoref{app:dca}}.
The leading product weights supervision at position $k$ by the acceptance overlaps of preceding drafts.

This soft weighting does not enforce the realized rejection boundary: positions after the first rejection can still receive nonzero gradients.
Since verification already marks the first-rejection position $j$ of each cycle, with $j=K{+}1$ when every draft is accepted, we introduce \emph{verify-gated masking} (VGM), which truncates the chain sum at and including the first rejected position of the recorded cycle,
\begin{equation}
    \mathcal{L}_{\mathrm{DCA}}^{\mathrm{VGM}} \;=\; -\log \sum_{l=1}^{\min(j,K)} \prod_{i=1}^{l} \alpha_i,
    \label{eq:vgm}
\end{equation}
removing supervision beyond the first rejection.
The chain sum includes the first rejected position, if any, and always includes the first draft position.
Supervision is therefore available from the first cycle, allowing training to begin even with a randomly initialized head.
The resulting supervision is also self-paced: its depth follows the realized accepted prefix, while gradient weights reflect the acceptance overlaps of preceding drafts.
This requires no shallow-depth warm-up or schedule for $K$.

Each update minimizes the mean of Eq.~\ref{eq:vgm} over cycles from the rollout policy and draft head, holding recorded tokens, target distributions, and $j$ fixed.
Gradients flow through reconstructed head distributions, excluding sampling and rejection decisions.
This fixed-record objective is an acceptance surrogate; unbiasedness for deployment-time expected accepted length is not assumed.

%% file: sections/experiments.tex
\section{Experiments}
\label{sec:experiments}

\subsection{Experimental Setup}

\noindent\textbf{Models.}
Qwen3-4B~\citep{yang2025qwen3} has no pretrained draft head: we attach a single randomly initialized EAGLE-style MTP layer and apply it recurrently at $K{=}5$.
Qwen3.5-4B-Base~\citep{qwen35blog} has a native MTP head ($K{=}3$), isolating the online training strategy without cold-start interference.
MiMo-7B-SFT~\citep{xiaomi2025mimo} has an MTP head ($K{=}1$), isolating how online training extends drafting depth from shallow capability.
The three heads take the same form, detailed in \textbf{Appendix~\ref{app:setup}}.
\textbf{Appendix~\ref{app:disc-models}} discusses which models \growmtp can accelerate and the signal is model-agnostic.

\noindent\textbf{Datasets.}
We evaluate \growmtp in two representative domains: mathematical reasoning and code reasoning.
For RL training, we use DAPO-Math-17K~\citep{yu2026dapo} and TACO-Verified~\citep{li2023taco}, two established datasets.
Inference evaluation uses AMC23~\citep{amc23}, AIME24, and AIME25~\citep{aime24,aime25} for mathematical reasoning, and LiveCodeBench v6~\citep{jain2025livecodebench} with its AtCoder and LeetCode slices for code reasoning, detailed in \textbf{Appendix~\ref{app:setup}}.
\growmtp thus requires no data constructed for draft-head training.

\noindent\textbf{Metrics.}
For the training evaluation, we measure rollout time, end-to-end step time with head updates included, and the acceptance length $\tau$, with speedups over the same run with the head frozen.
For the Inference evaluation, we report policy quality, Mean@16 for mathematical reasoning~\citep{team2025kimi} and Pass@4 for code reasoning~\citep{li2026scaling}, to assess the effect of draft-head training on policy quality, and the held-out $\tau$, with full protocols in \textbf{Appendix~\ref{app:setup-eval}}.

\noindent\textbf{Implementation Details.}
We implement \growmtp on verl~\citep{sheng2025hybridflow}, with rollouts served by SGLang~\citep{zheng2024sglang} and the draft head in its speculative decoding path.
Each RL step samples 64 prompts with 8 responses each, capped at 8192 tokens (16384 on Qwen3.5-4B-Base), and Qwen3-4B runs in thinking mode.
Rollouts and evaluation use rejection sampling at temperature 1.
Scaling $K$ introduces no new parameters, the same head applied recurrently, and the head is updated once per RL step from that step's recorded signals.
Full configurations are given in \textbf{Appendix~\ref{app:setup}}.

\subsection{RQ1: Can an RL Run Grow Its Own Draft Head from Scratch?}

\noindent\textbf{\growmtp vs. Baseline.}
We train Qwen3-4B for 500 RL steps in each domain with a randomly initialized $K{=}5$ head.
As shown in Table~\ref{tab:rq1-training} and \textbf{Appendix~\ref{app:results-rq1}}, the acceptance length $\tau$ grows from the autoregressive floor of 1.00 to 2.91 on mathematical reasoning and 2.64 on code reasoning, and the growth begins immediately, as the first draft position supplies valid supervision even under random initialization.
This growth converts into rollout speedups of $2.13\times$ and $1.93\times$, and with head-update cost included, end-to-end step speedups of $1.60\times$ and $1.56\times$.
Held-out benchmark scores are comparable.
The acceleration is not free early on, as the random head accepts little while already paying update cost, but per-step time falls below the baseline within the first 30 steps in both domains and subsequent steps remain faster.
\textbf{Appendix~\ref{app:disc-scale}} projects this speedup at longer rollouts and more steps.
\growmtp can thus train an MTP head from scratch during RL and use it to accelerate the run itself.
\textbf{Appendix~\ref{app:disc-serving}} shows that the grown head specializes to that run.

\input{tabs/rq1_training}

\input{tabs/rq1_online_offline}
\noindent\textbf{Online vs. Offline.}
We further compare against the established offline route, generating rollouts from the same initial checkpoint on the same dataset, training an identical head with the same objective, and then running RL with that fully trained head frozen throughout.
As shown in Table~\ref{tab:rq1-online-offline}, both routes reach nearly identical serving acceptance, 2.92 offline and 2.91 online.
The offline route, however, totals $555.3$ against $502.8$ GPU$\cdot$h for \growmtp, a $9.4\%$ difference over the 500-step run.
The gap arises at generation: offline training must produce its trajectories before any head exists, whereas \growmtp learns from rollouts that RL must generate anyway.
A separate offline stage is thus unnecessary for accelerating the run: \growmtp obtains an equally capable head at lower cost from the RL process itself.

\subsection{RQ2: How Should the Draft Head Be Trained Online?}

\noindent\textbf{Training Strategy.}
On Qwen3.5-4B-Base, we train for 200 RL steps in each domain under four strategies: frozen~\citep{chandiramani2026nemotron}, joint CE~\citep{wang2026joint}, detached CE~\citep{zeng2025glm}, and detached DCA, the last being \growmtp.
Frozen is the prevailing practice in current RL pipelines, and the two CE arms transplant the established recipe online, differing only in whether gradients reach the backbone.
As shown in Table~\ref{tab:rq2a-training} and \textbf{Appendix~\ref{app:results-rq2}}, joint training reaches the highest $\tau$, 3.99 and 3.42, yet its held-out quality is zero in both domains: $\tau$ can rise as the policy that produces it degrades.
The detached CE arm avoids the observed collapse and accelerates the run: $\tau$ reaches 3.18 and 3.06, and rollouts speed up by $1.29\times$ and $1.17\times$.
Under the same detachment at $K{=}3$, DCA leads CE on every training-side column and every held-out $\tau$, with end-to-end step speedups of $1.20\times$ and $1.12\times$ including head-update cost and held-out quality on par with the frozen head.
The draft head should thus be trained online, but with gradients detached from the policy and an objective aligned with acceptance: \growmtp further improves a pretrained head.

\vspace{-4pt}
\input{tabs/rq2a_training}

\noindent\textbf{Drafting Depth.}
We further evaluate how far \growmtp can extend the drafting depth, and which depth best serves RL acceleration.
We train on MiMo-7B-SFT at $K{=}3$, $5$, and $7$ for 200 RL steps in each domain, against the shipped head frozen at $K{=}1$.
As shown in Table~\ref{tab:rq2b-training} and \textbf{Appendix~\ref{app:results-rq2b}}, online training converts nominal depth into realized acceptance: at $K{=}5$, $\tau$ climbs to 4.04 and 3.72, beyond the structural ceiling of two for single-step drafting, rollouts accelerate by $1.93\times$ and $1.72\times$, and end-to-end steps by $1.41\times$ and $1.35\times$.
Greater depth, however, does not translate monotonically into end-to-end gains: at $K{=}7$, $\tau$ rises further to 4.38 and 4.00, yet step speedups fall to $1.30\times$ on math and $1.31\times$ on code, as both drafting and verification costs per cycle grow with depth while the marginal acceptance gain diminishes~\citep{chen2024sequoia}.
Held-out scores are comparable across depths.
The drafting depth should thus follow end-to-end acceleration rather than acceptance alone, with $K{=}5$ best in this setting: \growmtp also proves effective for a single-step head.

\input{tabs/rq2b_training}

\subsection{RQ3: How Does Each Component Contribute to Multi-Step Adaptation?}

Following the MiMo depth study, we isolate component effects using the same pretrained single-step head initialization. RQ1 separately establishes the complete \growmtp's from-scratch capability.

\noindent\textbf{Draft-Path Reconstruction.}
On MiMo-7B-SFT at $K{=}5$, we replace the reconstructed draft path with the committed response, teacher forcing under an unchanged objective~\citep{zhang2025learning}.
Teacher forcing supervises every position at every depth rather than one chain per cycle, and the two agree only at the first draft position.
As shown in Table~\ref{tab:rq3-dpr} and \textbf{Appendix~\ref{app:results-rq3}}, the added positions raise cost without raising aggregate acceptance length: acceptance ties within 0.03 while the step speedup falls from $1.41\times$ to $1.28\times$.
Resolved by depth, however, the discrepancy does not vanish: acceptance is indistinguishable at shallow positions, while the deepest retain a gap that narrows without closing.
Aggregate $\tau$ absorbs that residue, since the deepest positions contribute the smallest chain terms.
Thus, despite similar aggregate $\tau$ at $K{=}5$, teacher forcing retains a deeper-position gap and yields lower end-to-end efficiency than draft-path reconstruction.
\textbf{Appendix~\ref{app:tf}} analyzes this gap in detail.

\input{tabs/rq3_dpr}

\begin{figure}[b]
    \vspace{-0.4cm}
    \centering
    \begin{minipage}[c]{0.49\linewidth}\centering
        \includegraphics[height=10.5pt]{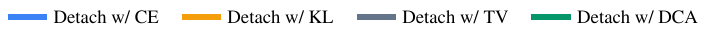}
    \end{minipage}\hfill
    \begin{minipage}[c]{0.49\linewidth}\centering
        \includegraphics[height=10.5pt]{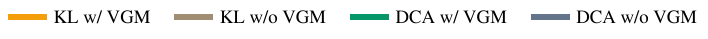}
    \end{minipage}\\[0.3em]
    \begin{subfigure}[c]{0.245\linewidth}
        \centering
        \includegraphics[width=\linewidth]{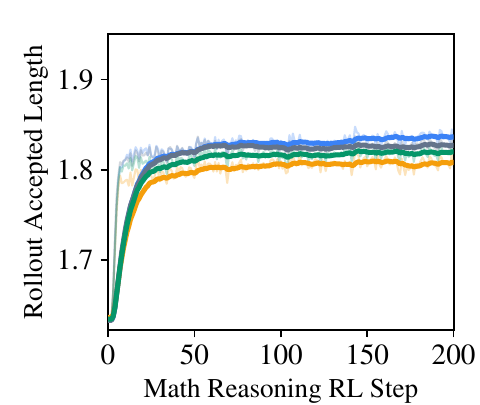}
        \caption{DCA ablation \textbf{(K=1)}.}
        \label{fig:rq3:d1-k1}
    \end{subfigure}\hfill
    \begin{subfigure}[c]{0.245\linewidth}
        \centering
        \includegraphics[width=\linewidth]{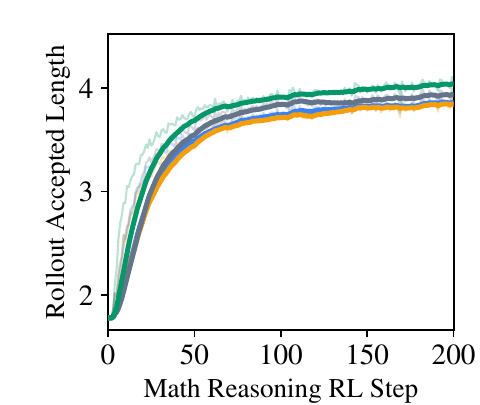}
        \caption{DCA ablation \textbf{(K=5)}.}
        \label{fig:rq3:d1-k5}
    \end{subfigure}\hfill
    \begin{subfigure}[c]{0.245\linewidth}
        \centering
        \includegraphics[width=\linewidth]{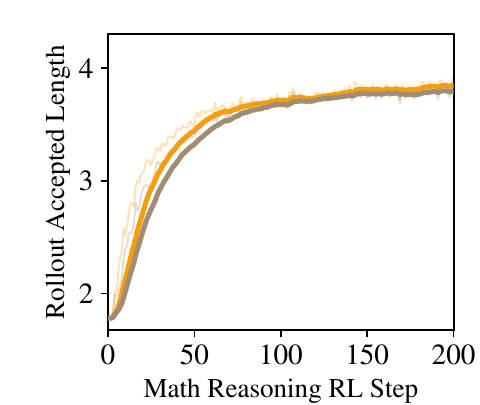}
        \caption{VGM ablation \textbf{(KL)}.}
        \label{fig:rq3:d2-kl}
    \end{subfigure}\hfill
    \begin{subfigure}[c]{0.245\linewidth}
        \centering
        \includegraphics[width=\linewidth]{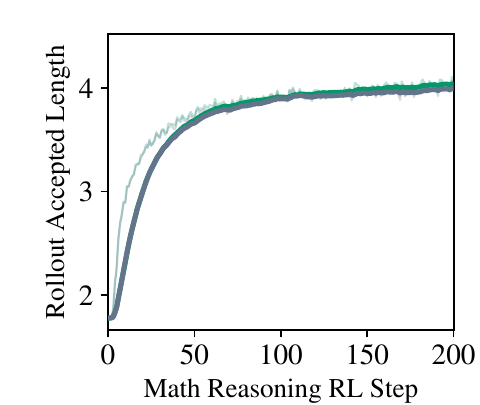}
        \caption{VGM ablation \textbf{(DCA)}.}
        \label{fig:rq3:d2-dca}
    \end{subfigure}
    \vspace{-4pt}
    \caption{\textbf{DCA and VGM support \growmtp from complementary sides (MiMo-7B-SFT).} (a,b) The $\tau$ gap of DCA over CE, KL, and TV opens with depth. (c,d) Removing VGM degrades $\tau$.}
    \label{fig:rq3}
\end{figure}

\noindent\textbf{DCA Loss.}
Retaining the setup of the depth study, we train in mathematical reasoning and replace only the objective: CE, KL~\citep{zhou2024distillspec}, and TV~\citep{li2026breaking}.
As shown in Figure~\ref{fig:rq3}(a,b) and \textbf{Appendix~\ref{app:results-rq3}}, at $K{=}1$ the four objectives differ by less than 0.03 in $\tau$, as the chain reduces to a single acceptance rate leaving no depth structure.
At $K{=}5$, DCA leads consistently: $\tau$ reaches 4.04 against 3.86 for CE, 3.85 for KL, and 3.94 for TV, and the end-to-end step speedup $1.41\times$ against $1.36\times$ to $1.37\times$, head-update cost included.
Relative to CE and KL, the advantage widens with depth: DCA is built directly on the acceptance chain, whereas per-position imitation ignores the chain dependence.
TV and DCA use the same acceptance-chain surrogate; the logarithm changes their relative cycle weights in the batch average, thus favoring cycles with smaller chain values.

\noindent\textbf{Verify-Gated Masking.}
We remove VGM under both the KL and DCA objectives.
As shown in Figure~\ref{fig:rq3}(c,d) and \textbf{Appendix~\ref{app:results-rq3}}, both degrade: the step speedup of KL falls from $1.37\times$ to $1.28\times$, showing an end-to-end efficiency benefit from masking post-rejection positions.
DCA degrades less, from $1.41\times$ to $1.35\times$. Its soft weighting can retain post-rejection gradients; VGM enforces the rejection boundary and provides an additional efficiency gain.
Together, the components improve multi-step adaptation: reconstruction lowers training cost at comparable acceptance length, DCA improves acceptance, and VGM further improves acceptance and end-to-end efficiency.

%% file: tabs/rq1_training.tex
\begin{table}[t]
    \centering
    \small
    \setlength{\tabcolsep}{4pt}
    \caption{\textbf{Training evaluation on Qwen3-4B (500 RL steps, $K{=}5$).} A draft head grown from scratch delivers end-to-end step speedups of $1.60\times$ and $1.56\times$ over autoregressive decoding ($K{=}0$).}
    \label{tab:rq1-training}
    \vspace{-4pt}
    \resizebox{\linewidth}{!}{%
    \begin{tabular}{@{}llcccccccccc@{}}
        \toprule
        & & \multicolumn{5}{c}{\textbf{Math Reasoning RL}} & \multicolumn{5}{c}{\textbf{Code Reasoning RL}} \\
        \cmidrule(lr){3-7} \cmidrule(lr){8-12}
        Method & $K$ & Rollout (s) & Speedup & Step (s) & Speedup & $\tau$ & Rollout (s) & Speedup & Step (s) & Speedup & $\tau$ \\
        \midrule
        AR RL              & $0$ & $523.86$ & $1.00\times$ & $723.51$ & $1.00\times$ & $1.00$ & $771.22$ & $1.00\times$ & $1028.39$ & $1.00\times$ & $1.00$ \\
        \rowcolor{brandgreen}
        \textbf{\growmtp} & $\mathbf{5}$ & $\mathbf{245.95}$ & $\mathbf{2.13\times}$ & $\mathbf{452.53}$ & $\mathbf{1.60\times}$ & $\mathbf{2.91}$ & $\mathbf{398.65}$ & $\mathbf{1.93\times}$ & $\mathbf{659.39}$ & $\mathbf{1.56\times}$ & $\mathbf{2.64}$ \\
        \bottomrule
    \end{tabular}%
    }
    \vspace{-0.4cm}
\end{table}

%% file: tabs/rq1_online_offline.tex
\begin{wraptable}{r}{0.50\linewidth}
    \centering
    \small
    \setlength{\tabcolsep}{3pt}
    \caption{\textbf{\growmtp avoids 161.6 GPU$\cdot$h offline head-training stage.} Costs in GPU$\cdot$h; Qwen3-4B, 500 RL steps on DAPO-Math-17K, $K{=}5$.}
    \label{tab:rq1-online-offline}
    \begin{tabular}{@{}lcccc@{}}
        \toprule
        Method & Offline (h)& RL (h)& Total (h)& $\tau$ \\
        \midrule
        AR RL (K = 0)     & $0$     & $803.9$ & $803.9$          & $1.00$ \\
        Offline (K = 5)  & $161.6$ & $393.7$ & $555.3$          & $2.92$ \\
        \rowcolor{brandgreen}
        \growmtp (K = 5) & $0$     & $502.8$ & $\mathbf{502.8}$ & $2.91$ \\
        \bottomrule
    \end{tabular}
\end{wraptable}

%% file: tabs/rq2a_training.tex
\begin{table}[htbp]
    \centering
    \small
    \setlength{\tabcolsep}{4pt}
    \caption{\textbf{Training evaluation on Qwen3.5-4B-Base (200 RL steps, $K{=}3$).} Both detached arms accelerate. DCA leads among quality-preserving strategies. Joint CE speeds up code but scores zero.}
    \label{tab:rq2a-training}
    \vspace{-4pt}
    \resizebox{\linewidth}{!}{%
    \begin{tabular}{@{}llcccccccccc@{}}
        \toprule
        & & \multicolumn{5}{c}{\textbf{Math Reasoning RL}} & \multicolumn{5}{c}{\textbf{Code Reasoning RL}} \\
        \cmidrule(lr){3-7} \cmidrule(lr){8-12}
        Training Mode & Loss & Rollout (s) & Speedup & Step (s) & Speedup & $\tau$ & Rollout (s) & Speedup & Step (s) & Speedup & $\tau$ \\
        \midrule
        Frozen   & --  & $357.62$ & $1.00\times$ & $554.42$ & $1.00\times$ & $2.65$ & $500.09$ & $1.00\times$ & $778.54$ & $1.00\times$ & $2.59$ \\
        Joint    & CE  & $372.87$ & $0.96\times$ & $643.76$ & $0.86\times$ & $3.99$ & $258.34$ & $1.94\times$ & $411.88$ & $1.89\times$ & $3.42$ \\
        Detached & CE  & $277.14$ & $1.29\times$ & $507.91$ & $1.09\times$ & $3.18$ & $426.88$ & $1.17\times$ & $719.40$ & $1.08\times$ & $3.06$ \\
        \rowcolor{brandgreen}
        \textbf{Detached} & \textbf{DCA (\growmtp)} & $\mathbf{263.86}$ & $\mathbf{1.36\times}$ & $\mathbf{461.56}$ & $\mathbf{1.20\times}$ & $\mathbf{3.22}$ & $\mathbf{408.89}$ & $\mathbf{1.22\times}$ & $\mathbf{694.00}$ & $\mathbf{1.12\times}$ & $\mathbf{3.11}$ \\
        \bottomrule
    \end{tabular}%
    }
    \vspace{-0.2cm}
\end{table}

%% file: tabs/rq2b_training.tex
\begin{table}[t]
    \centering
    \small
    \setlength{\tabcolsep}{4pt}
    \caption{\textbf{Training evaluation across drafting depths on MiMo-7B-SFT (200 RL steps).} $\tau$ rises monotonically with depth, while end-to-end step speedups peak at $K{=}5$, the highlighted depth.}
    \label{tab:rq2b-training}
    \vspace{-4pt}
    \resizebox{\linewidth}{!}{%
    \begin{tabular}{@{}llcccccccccc@{}}
        \toprule
        & & \multicolumn{5}{c}{\textbf{Math Reasoning RL}} & \multicolumn{5}{c}{\textbf{Code Reasoning RL}} \\
        \cmidrule(lr){3-7} \cmidrule(lr){8-12}
        Method & $K$ & Rollout (s) & Speedup & Step (s) & Speedup & $\tau$ & Rollout (s) & Speedup & Step (s) & Speedup & $\tau$ \\
        \midrule
        Frozen    & $1$ & $318.66$ & $1.00\times$ & $523.02$ & $1.00\times$ & $1.65$ & $441.81$ & $1.00\times$ & $707.42$ & $1.00\times$ & $1.59$ \\
        \growmtp & $3$ & $176.26$ & $1.81\times$ & $378.88$ & $1.38\times$ & $3.25$ & $264.69$ & $1.67\times$ & $528.64$ & $1.34\times$ & $3.11$ \\
        \rowcolor{brandgreen}
        \growmtp & $5$ & $164.84$ & $1.93\times$ & $\mathbf{371.02}$ & $\mathbf{1.41\times}$ & $4.04$ & $256.86$ & $1.72\times$ & $\mathbf{525.31}$ & $\mathbf{1.35\times}$ & $3.72$ \\
        \growmtp & $7$ & $171.24$ & $1.86\times$ & $402.20$ & $1.30\times$ & $4.38$ & $252.03$ & $1.75\times$ & $542.02$ & $1.31\times$ & $4.00$ \\
        \bottomrule
    \end{tabular}%
    }
    \vspace{-0.4cm}
\end{table}

%% file: tabs/rq3_dpr.tex
\begin{table}[htbp]
    \centering
    \footnotesize
    \setlength{\tabcolsep}{3pt}
    \caption{\textbf{Ablation of Draft-Path Reconstruction on MiMo-7B-SFT (200 RL steps, $K{=}5$).} Acceptance ties within $0.03$, and reconstruction lifts the end-to-end step speedup from $1.28\times$ to $1.41\times$.}
    \label{tab:rq3-dpr}
    \vspace{-4pt}
    \resizebox{\linewidth}{!}{%
    \begin{tabular}{@{}lccccccccccc@{}}
        \toprule
        & \multicolumn{5}{c}{\textbf{Math Reasoning RL}} & \multicolumn{6}{c}{\textbf{Math Reasoning Eval}} \\
        \cmidrule(lr){2-6} \cmidrule(lr){7-12}
        & \multicolumn{5}{c}{DAPO-Math-17K} & \multicolumn{2}{c}{AMC23} & \multicolumn{2}{c}{AIME24} & \multicolumn{2}{c}{AIME25} \\
        \cmidrule(lr){2-6} \cmidrule(lr){7-8} \cmidrule(lr){9-10} \cmidrule(lr){11-12}
        Method & Rollout (s) & Speedup & Step (s) & Speedup & $\tau$ & Mean@16 & $\tau$ & Mean@16 & $\tau$ & Mean@16 & $\tau$ \\
        \midrule
        Teacher forcing & $169.05$ & $1.89\times$ & $408.09$ & $1.28\times$ & $4.02$ & $91.56$ & $3.93$ & $54.37$ & $3.82$ & $43.96$ & $3.79$ \\
        \rowcolor{brandgreen}
        \growmtp & $\mathbf{164.84}$ & $\mathbf{1.93\times}$ & $\mathbf{371.02}$ & $\mathbf{1.41\times}$ & $\mathbf{4.04}$ & $91.25$ & $\mathbf{3.95}$ & $56.87$ & $\mathbf{3.84}$ & $46.25$ & $\mathbf{3.82}$ \\
        \bottomrule
    \end{tabular}%
    }
\end{table}

%% file: sections/related_work.tex
\section{Related Work}
\label{sec:related}

\subsection{Speculative Decoding}

The draft model has evolved from independent small models~\citep{miao2024specinfer} to lightweight draft heads that reuse the backbone's hidden states~\citep{cai2024medusa,li2024eagle}, now common components of frontier models~\citep{liu2024deepseek,qwen35blog,xiaomi2025mimo}.
Draft heads can be trained jointly with the backbone or through offline distillation from target-generated data~\citep{gloeckle2024better,li2026eagle,hu2026bridging}, or adapted online to the served query distribution~\citep{liu2023online}.
HASS~\citep{zhang2025learning} and EAGLE-3~\citep{li2026eagle} seek to align draft-head training states with those encountered during inference.
After SFT warm-up, Draft-OPD~\citep{lei2026draft} replays draft paths in a separate online distillation stage and treats accepted and rejected positions differently.
\growmtp shares this alignment motivation, reconstructing training states from draft paths and matching verification distributions recorded during RL.
For acceptance-oriented training, Bebop~\citep{li2026breaking} introduces a multi-step TV-chain objective.
DCA applies a logarithmic transformation to this objective to adjust the relative gradient weights of cycles and combines it with VGM, which retains supervision up to and including the first rejected position.

\subsection{Accelerating Reinforcement Learning for LLMs}

Recent work accelerates the rollout stage that dominates RL wall-clock.
One line integrates high-throughput inference engines into the RL training loop to reduce rollout cost~\citep{hu2025openrlhf,shen2024nemo}.
For example, VeRL~\citep{sheng2025hybridflow} adopts a hybrid controller architecture that decouples RL control flow from distributed computation.
Another line optimizes rollout scheduling to reduce tail waiting~\citep{noukhovitch2025asynchronous}.
For example, AReaL~\citep{fu2026areal} uses asynchronous execution to reduce synchronization stalls.
Speculative decoding increases the number of tokens committed per verification step: RhymeRL~\citep{he2025history} constructs drafts from historical rollouts, while ReSpec~\citep{chen2026respec} updates draft models with RL supervision and includes an offline warm-up stage.
\growmtp supports random initialization, integrating drafting, verification, and head updates into the same RL run without a separate offline training stage.
We measure end-to-end speedups over the full run, including training cost, and evaluate it on pretrained heads.

%% file: sections/conclusion.tex
\section{Conclusion}
\label{sec:conclusion}

In this paper, we propose \growmtp, a from-scratch online draft-head training method for accelerating the rollout stage of RL training.
We observe that RL training itself provides both conditions required for online draft-head training: its rollout distribution is far narrower than that of pretraining, and its verification step continuously produces supervision signals aligned with this distribution.
Building on these observations, \growmtp trains the draft head online within the RL loop using this signal directly, enabling a randomly initialized head to grow entirely within RL without any offline pretraining or warm-up.
Comprehensive evaluations on Qwen3-4B (no draft head), MiMo-7B-SFT (weak head), and Qwen3.5-4B-Base (strong head) show that \growmtp achieves rollout speedups of $2.13\times$, $1.93\times$, and $1.36\times$, and end-to-end speedups of $1.60\times$, $1.41\times$, and $1.20\times$, respectively, with comparable held-out benchmark scores in the reported runs.
These results demonstrate that speculative decoding for RL acceleration no longer requires a separate offline training stage.
\growmtp therefore can be integrated into existing RL training frameworks as a modular acceleration component, particularly offering a from-scratch acceleration path for models without pretrained MTP heads.

%% file: sections/appendix.tex
\clearpage
\section{Notation and Definitions}
\label{app:notation}
Table~\ref{tab:notation} summarizes the notation used throughout the paper.
\input{tabs/notations}

\section{Experimental Setup and Implementation}
\label{app:setup}

\subsection{Models and Draft Heads}
\label{app:setup-models}

\noindent\textbf{Baseline Models.}
We evaluate \growmtp on three models: Qwen3-4B, Qwen3.5-4B-Base, and MiMo-7B-SFT.
The three differ in the draft head they ship: Qwen3-4B has never shipped one, while Qwen3.5-4B-Base and MiMo-7B-SFT each ship a native head.
The drafting depth of such a head is not fixed by its architecture, so we adopt the depth recommended by each model's release: $K{=}3$ for Qwen3.5-4B-Base and $K{=}1$ for MiMo-7B-SFT.
Apart from this shipped state, the draft heads of the three models take the same single-layer recurrent MTP form throughout this paper, detailed below.

\noindent\textbf{MTP Architecture.}
All three draft heads take the EAGLE-style single-layer recurrent MTP form.
The module contains a single decoder layer.
It drafts one position at a time, consuming the hidden state and the token embedding of the position before it: the two are normalized by separate RMSNorms, concatenated, and projected back to the hidden dimension by a linear layer without bias before entering the decoder layer.
It maintains a KV cache of its own, which grows by one entry with every position it drafts.
The head contains neither an embedding nor an output projection and takes both from the backbone.
The heads of the three models therefore run through the same drafting and training implementation, and the family difference is confined to the type of the decoder layer, one each from Qwen3, Qwen3.5, and Qwen2.
The native head of MiMo concatenates the two inputs in the opposite order, and weight loading swaps the two halves of that linear layer accordingly.

\noindent\textbf{From-Scratch Initialization.}
The draft head of Qwen3-4B is built in this same form, identical to the two native heads down to the parameter names.
Randomization covers this newly built module alone: every linear layer within it is drawn from $\mathcal{N}(0,\, 0.02)$, the value of the backbone's own \texttt{initializer\_range}, and every RMSNorm weight is set to one.
The head is materialized once into a checkpoint under seed 0, and the training side and the SGLang drafting side load the same weights from it, so no divergence between two independent initializations arises.
The embedding and the output head are inherited from the backbone rather than randomized, so training from scratch refers to the drafting capability and not to every parameter that participates in drafting.

\subsection{RL Training Configuration}
\label{app:setup-rl}

\noindent\textbf{Policy Optimization.}
Policy optimization uses GRPO with one modification: the negative-advantage branch carries a weight of $0.5$, and every other term follows the standard form, dual clipping and token-level averaging among them.
The objective retains a KL regularizer at coefficient $0.01$, whose reference policy is the initial checkpoint and stays fixed throughout training.
The policy is updated by AdamW in verl, at a constant learning rate of $1 \times 10^{-6}$ and with a clipping range of $0.2$.
All three models, both domains, and every ablation arm share this configuration, so the differences between groups arise only from the variable isolated in each comparison with policy optimization held fixed.

\noindent\textbf{Training Data.}
Mathematical reasoning uses DAPO-Math-17K and code reasoning uses TACO-Verified.
The reward comes from matching the reference answer in mathematical reasoning and from matching the standard output of the generated program in code reasoning.
Both datasets are used as released, without difficulty filtering or subsampling, with head supervision drawn from RL rollouts.

\noindent\textbf{Rollout and Batching.}
All runs execute on a single node of $8\times$H800 GPUs, with the model sharded across the eight devices.
Each step samples 64 prompts and generates 8 responses per prompt, and the policy update uses a mini-batch of 64.
Prompts are truncated to 1024 tokens, and responses are capped at 8192 tokens on Qwen3-4B and MiMo-7B-SFT and at 16384 tokens on Qwen3.5-4B-Base.
Rollouts are served by SGLang at temperature $1.0$ and top-$p$ $1.0$, with draft tokens accepted by rejection sampling.
Qwen3-4B runs in thinking mode, and the other two models are not thinking models.
All runs fix the data order with seed 1, keeping the data order consistent across methods.

\clearpage
\subsection{Draft-Head Training Configuration}
\label{app:setup-head}

\noindent\textbf{Draft Head Training.}
In every \growmtp run the head is trained under the DCA objective with detached gradients and with VGM enabled throughout.
The ablation arms each change one of these: the objective becomes CE, KL, or the TV-chain objective of \autoref{app:dca}, the gradients become joint or frozen, or VGM is removed.
The drafting depth is $K{=}5$ on Qwen3-4B, $K{=}3$ on Qwen3.5-4B-Base, and $K{=}3$, $5$, and $7$ on MiMo-7B-SFT, whose frozen baseline runs at $K{=}1$.
The head is updated once per RL step, from the verification signals recorded in that step's own rollouts.
Training includes no freeze phase, and the head is updated from the first step to the last using online RL supervision.

\noindent\textbf{Optimizer and Learning Rate.}
A shared AdamW optimizer updates the head and backbone in separate parameter groups with independent learning rates.
Head learning rates are $3 \times 10^{-4}$ for random initialization and $1 \times 10^{-4}$ for the two pretrained heads.
Both head rates warm up over 10 steps, then follow a cosine decay to one tenth of their peak; the policy learning rate stays constant.

\subsection{Evaluation Protocol}
\label{app:setup-eval}

\noindent\textbf{Training Metrics.}
On the training side we report two times, the rollout time and the end-to-end step time, the latter including the cost of head updates.
Both are per-step means over the whole of training.
The denominator of a speedup is the run with the head frozen under the same script, the same data, and the same hyperparameters, and on Qwen3-4B that denominator carries no head and reduces to autoregressive decoding.
The training-side $\tau$ is instead averaged over the final 10 steps: the times are costs incurred throughout, whereas $\tau$ is the capability of the head once it has converged in the run.

\noindent\textbf{Inference Evaluation Benchmarks.}
Mathematical reasoning is evaluated on AMC23, AIME24, and AIME25, with 40, 30, and 30 questions respectively.
Code reasoning is evaluated on LiveCodeBench v6, whose 1055 questions include the AtCoder and LeetCode platform slices at 602 and 444 questions.
Judging uses the same string-matching verifier as training in mathematical reasoning and the official LiveCodeBench framework in code reasoning, with a timeout of 6 seconds for each separate test case.

\noindent\textbf{Evaluation Configuration.}
SGLang generates responses at temperature $1.0$ and top-$p$ $0.7$, capped at 16384 tokens.
Speculative decoding uses rejection sampling, as in training.
Evaluation uses each run's trained depth, each frozen baseline's shipped depth, and no draft head for the AR baseline.

\noindent\textbf{Evaluation Metrics.}
Policy quality is reported as Mean@16 in mathematical reasoning, the mean accuracy over 16 samples.
In code reasoning it is reported as Pass@4, the fraction of questions on which at least one of 4 attempts passes every test.
Drafting capability is the held-out $\tau$, one plus the ratio of accepted tokens to verifications, measured on the same generations used to assess quality.

\clearpage
\section{Additional Results}
\label{app:results}

\subsection{Additional Results for Growing from Scratch (RQ1)}
\label{app:results-rq1}

\begin{figure}[h]
    \centering
    \begin{subfigure}[c]{0.40\linewidth}
        \centering
        \includegraphics[width=\linewidth]{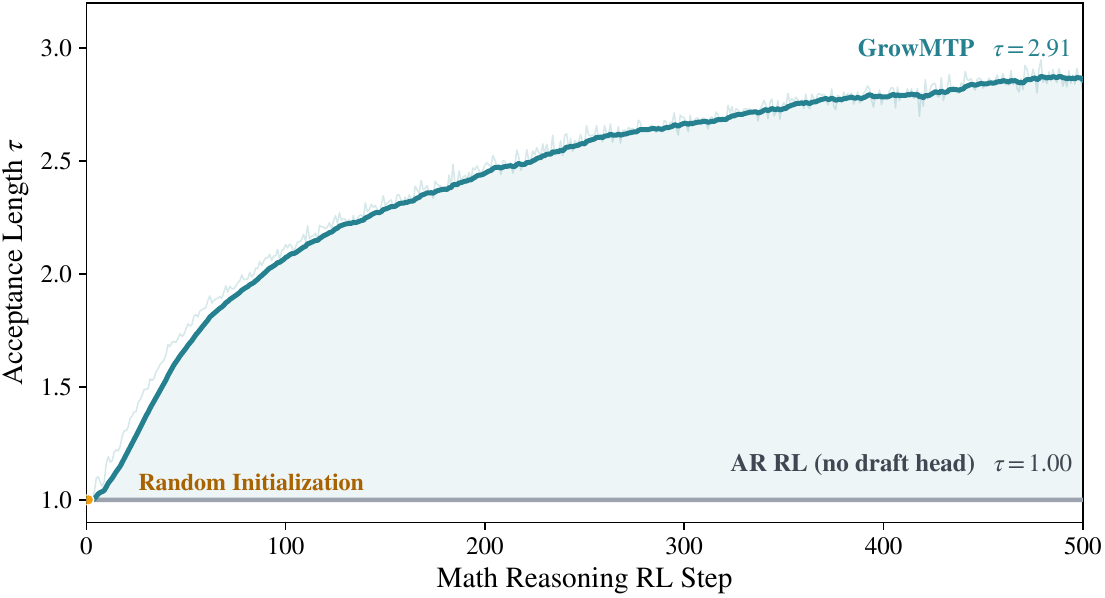}
        \caption{Acceptance length $\tau$.}
        \label{fig:rq1-math:eal}
    \end{subfigure}
    \hspace{0.066\linewidth}
    \begin{subfigure}[c]{0.40\linewidth}
        \centering
        \includegraphics[width=\linewidth]{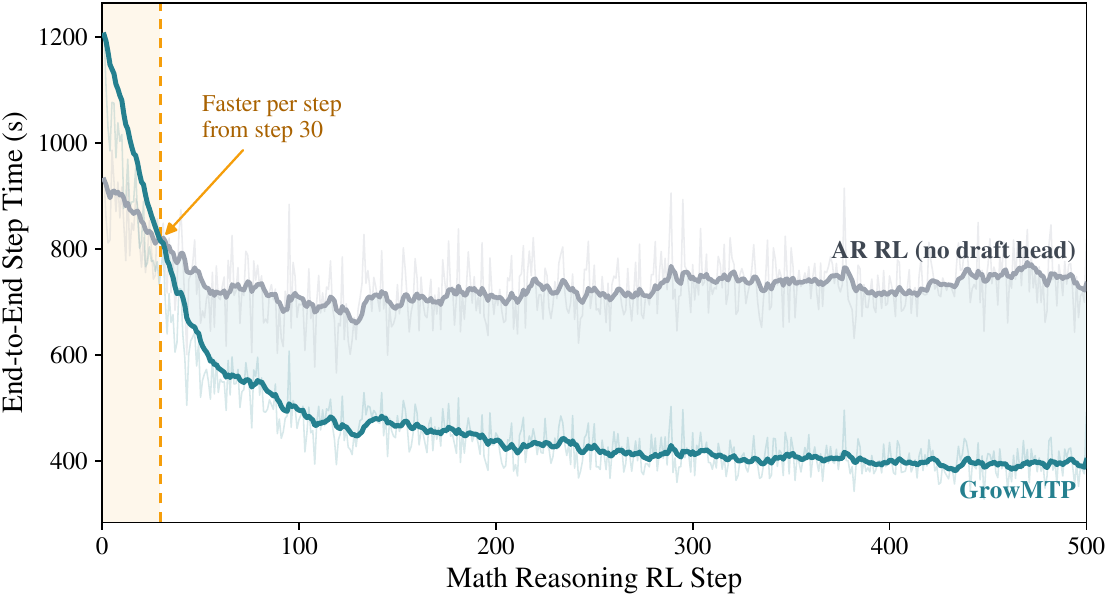}
        \caption{End-to-end step time.}
        \label{fig:rq1-math:time}
    \end{subfigure}
    \vspace{-4pt}
    \caption{\textbf{Math reasoning (Qwen3-4B): a draft head grows from scratch within RL and accelerates the run itself.} $\tau$ rises to $2.91$, and step time drops below the no-head baseline within $30$ steps.}
    \label{fig:rq1-math}
    \vspace{-6pt}
\end{figure}

\noindent\textbf{Math Reasoning RL Training.}
Figure~\ref{fig:rq1-math} shows the mathematical reasoning run: the acceptance length $\tau$ and the end-to-end step time over the 500 RL steps.
From random initialization at the autoregressive floor, $\tau$ rises steepest over the first hundred steps and is still climbing at step 500.
Step time opens above the no-head baseline, the early overhead visible on the curve, and crosses below it at step 30: every later step runs faster, and the widening gap tracks the growing $\tau$ throughout the run.

\begin{figure}[h]
    \centering
    \begin{subfigure}[c]{0.40\linewidth}
        \centering
        \includegraphics[width=\linewidth]{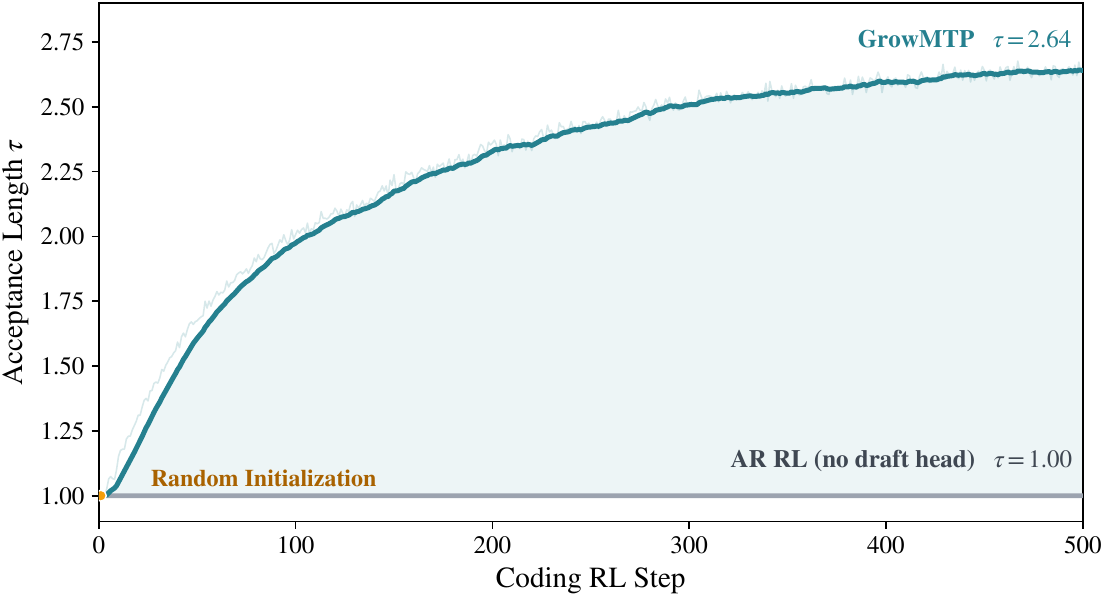}
        \caption{Acceptance length $\tau$.}
        \label{fig:rq1-code:eal}
    \end{subfigure}
    \hspace{0.066\linewidth}
    \begin{subfigure}[c]{0.40\linewidth}
        \centering
        \includegraphics[width=\linewidth]{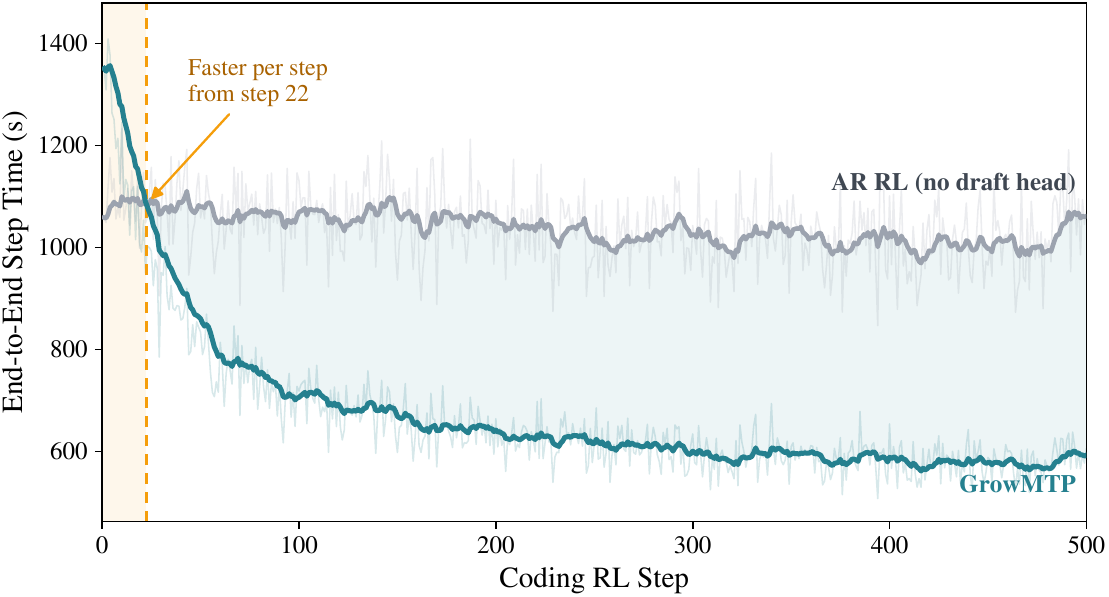}
        \caption{End-to-end step time.}
        \label{fig:rq1-code:time}
    \end{subfigure}
    \vspace{-4pt}
    \caption{\textbf{Code reasoning (Qwen3-4B): the from-scratch growth pattern of Figure~\ref{fig:rq1-math} holds.} $\tau$ rises to $2.64$ over $500$ RL steps, and step time drops below the no-head baseline within $22$ steps.}
    \label{fig:rq1-code}
    \vspace{-6pt}
\end{figure}

\noindent\textbf{Code Reasoning RL Training.}
Figure~\ref{fig:rq1-code} shows the training trajectories of the code reasoning run: $\tau$ and the end-to-end step time over the 500 RL steps.
The trajectories mirror the mathematical reasoning run of Figure~\ref{fig:rq1-math}: $\tau$ leaves the autoregressive floor immediately and climbs to 2.64 over the 500 steps, and step time falls below the no-head baseline within 22 steps.
The from-scratch growth pattern is thus not specific to a single training domain: both domains support online head training.

\noindent\textbf{Inference Evaluation.}
Table~\ref{tab:rq1-eval} reports the held-out evaluation of the two runs, pairing policy quality with $\tau$ on each benchmark.
Policy quality stays on par with the autoregressive baseline on all six benchmarks, with Mean@16 and Pass@4 differing by at most 2.29 points.
The trained heads reach $\tau$ of 2.30--2.34 on the mathematical benchmarks and 2.11--2.27 on the code benchmarks: the drafting capability learned on training rollouts carries over to held-out evaluation prompts of the same domain.

\input{tabs/rq1_eval}

\clearpage
\subsection{Additional Results for Training Strategy (RQ2)}
\label{app:results-rq2}

\begin{figure}[h]
    \centering
    \includegraphics[width=0.7\linewidth]{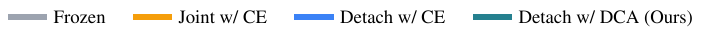}\\[0.1em]
    \begin{subfigure}[c]{0.245\linewidth}
        \centering
        \includegraphics[width=\linewidth]{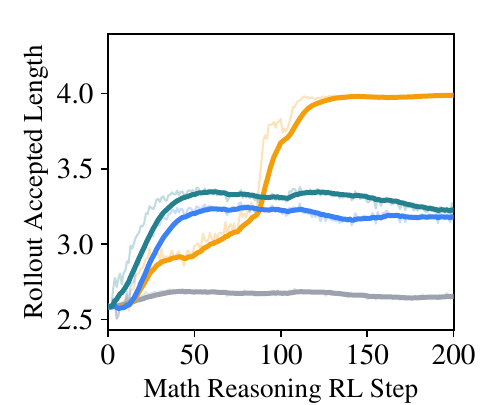}
        \caption{Acceptance length.}
        \label{fig:rq2a-math:eal}
    \end{subfigure}\hfill
    \begin{subfigure}[c]{0.245\linewidth}
        \centering
        \includegraphics[width=\linewidth]{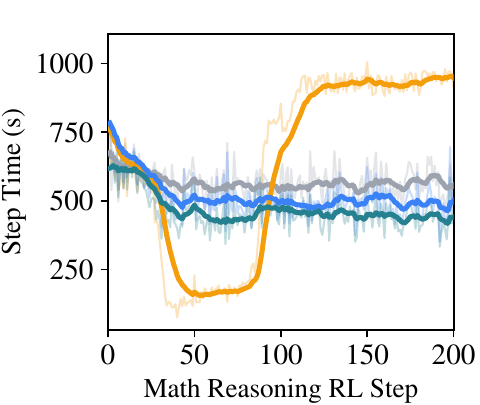}
        \caption{End-to-end step time.}
        \label{fig:rq2a-math:time}
    \end{subfigure}\hfill
    \begin{subfigure}[c]{0.245\linewidth}
        \centering
        \includegraphics[width=\linewidth]{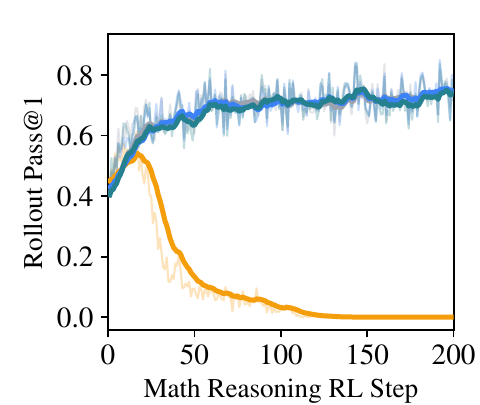}
        \caption{Pass@1.}
        \label{fig:rq2a-math:pass1}
    \end{subfigure}\hfill
    \begin{subfigure}[c]{0.245\linewidth}
        \centering
        \includegraphics[width=\linewidth]{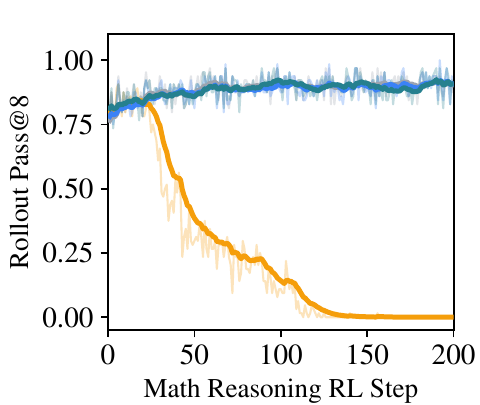}
        \caption{Pass@8.}
        \label{fig:rq2a-math:pass8}
    \end{subfigure}
    \vspace{-6pt}
    \caption{\textbf{Math reasoning (Qwen3.5-4B-Base): four training strategies.} Both detached arms raise $\tau$ above the frozen head and DCA leads, while joint CE reaches the highest $\tau$ at zero held-out quality.}
    \label{fig:rq2a-math}
    \vspace{-6pt}
\end{figure}

\noindent\textbf{Math Reasoning RL.}
Figure~\ref{fig:rq2a-math} shows the four training strategies on mathematical reasoning: $\tau$, the end-to-end step time, and held-out quality (Pass@1 and Pass@8) over the 200 RL steps.
Both detached arms raise $\tau$ above the frozen head while their quality curves stay level throughout.
Joint CE traces the opposite pattern: its $\tau$ climbs to the highest of the four while its Pass@1 and Pass@8 collapse to zero, the covariation behind reading a rising $\tau$ as no evidence of a healthy RL training run.

\begin{figure}[h]
    \centering
    \includegraphics[width=0.7\linewidth]{figures/Qwen3.5-4B-Base/legend_groupA.pdf}\\[0.1em]
    \begin{subfigure}[c]{0.245\linewidth}
        \centering
        \includegraphics[width=\linewidth]{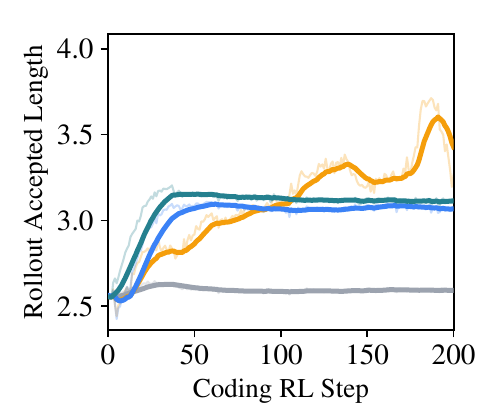}
        \caption{Acceptance length.}
        \label{fig:rq2a-code:eal}
    \end{subfigure}\hfill
    \begin{subfigure}[c]{0.245\linewidth}
        \centering
        \includegraphics[width=\linewidth]{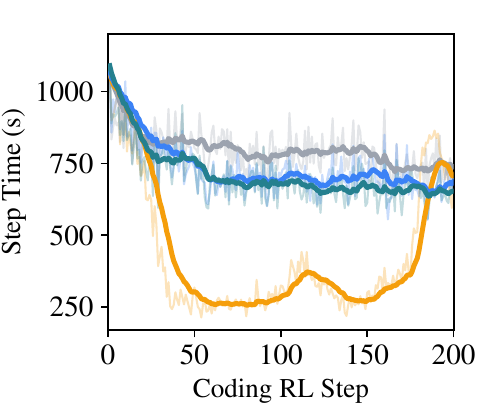}
        \caption{End-to-end step time.}
        \label{fig:rq2a-code:time}
    \end{subfigure}\hfill
    \begin{subfigure}[c]{0.245\linewidth}
        \centering
        \includegraphics[width=\linewidth]{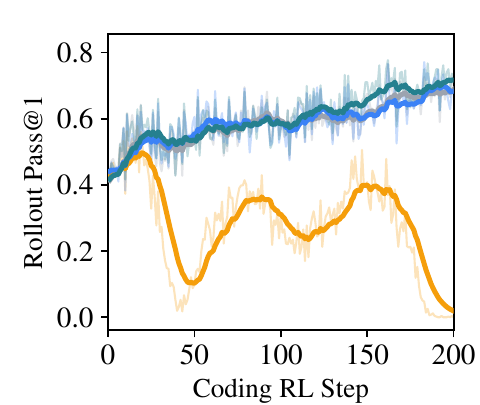}
        \caption{Pass@1.}
        \label{fig:rq2a-code:pass1}
    \end{subfigure}\hfill
    \begin{subfigure}[c]{0.245\linewidth}
        \centering
        \includegraphics[width=\linewidth]{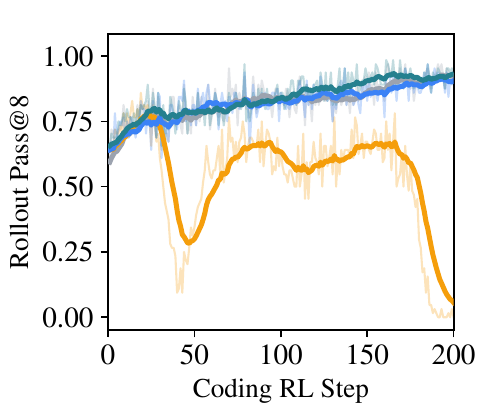}
        \caption{Pass@8.}
        \label{fig:rq2a-code:pass8}
    \end{subfigure}
    \vspace{-6pt}
    \caption{\textbf{Code reasoning (Qwen3.5-4B-Base): the same comparison.} The quality-preserving ranking mirrors the math domain, while joint CE reaches the largest speedups at zero held-out quality.}
    \label{fig:rq2a-code}
    \vspace{-6pt}
\end{figure}

\noindent\textbf{Code Reasoning RL.}
Figure~\ref{fig:rq2a-code} repeats the comparison on code reasoning.
Among the quality-preserving strategies the trajectories mirror the math domain: both detached arms accelerate the run with level quality curves, and DCA leads.
Joint CE, however, reaches the lowest step time of the four here after slowing the run on math, at zero quality in both domains: the speed of a collapsed policy flips sign across domains and cannot establish whether RL accelerates with preserved policy quality.

\noindent\textbf{Inference Evaluation.}
Table~\ref{tab:rq2a-eval} reports the held-out evaluation of the four strategies, pairing policy quality with $\tau$ on each benchmark.
Among the quality-preserving strategies, DCA reaches $\tau$ of 3.22--3.31 against 3.10--3.19 for detached CE, leading on all six benchmarks.
Joint CE scores zero on every benchmark in both domains, and its $\tau$ of 3.98--3.99 is measured on this collapsed policy: these high drafting numbers are excluded from the comparison among the quality-preserving strategies.

\input{tabs/rq2a_eval}

\subsection{Additional Results for Drafting Depth (RQ2)}
\label{app:results-rq2b}

\begin{figure}[h]
    \centering
    \includegraphics[width=0.7\linewidth]{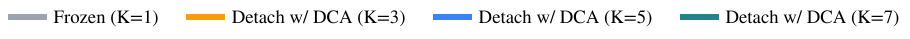}\\[0.3em]
    \begin{subfigure}[c]{0.245\linewidth}
        \centering
        \includegraphics[width=\linewidth]{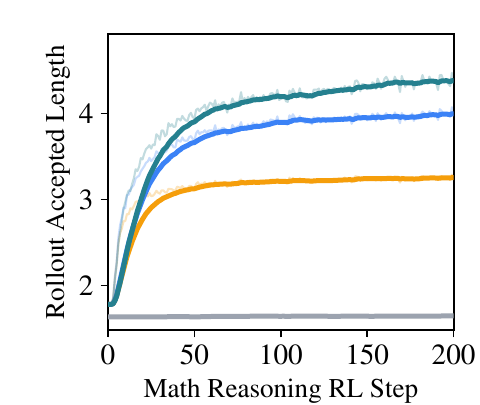}
        \caption{$\tau$ \textbf{(Math)}.}
        \label{fig:rq2b:eal-math}
    \end{subfigure}\hfill
    \begin{subfigure}[c]{0.245\linewidth}
        \centering
        \includegraphics[width=\linewidth]{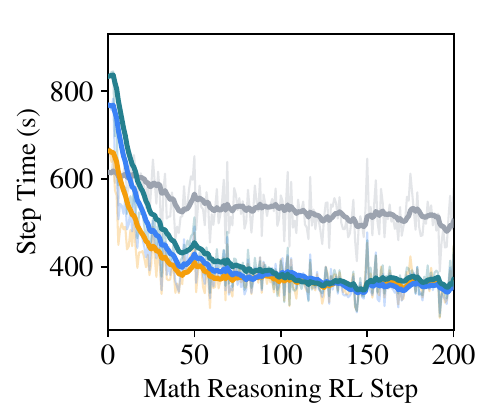}
        \caption{Step time \textbf{(Math)}.}
        \label{fig:rq2b:time-math}
    \end{subfigure}\hfill
    \begin{subfigure}[c]{0.245\linewidth}
        \centering
        \includegraphics[width=\linewidth]{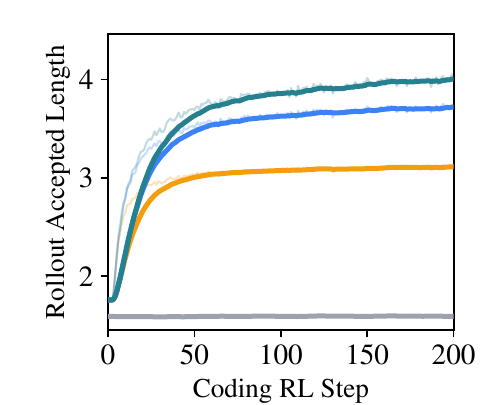}
        \caption{$\tau$ \textbf{(Code)}.}
        \label{fig:rq2b:eal-code}
    \end{subfigure}\hfill
    \begin{subfigure}[c]{0.245\linewidth}
        \centering
        \includegraphics[width=\linewidth]{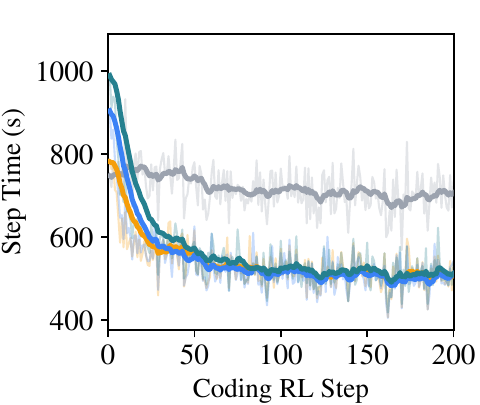}
        \caption{Step time \textbf{(Code)}.}
        \label{fig:rq2b:time-code}
    \end{subfigure}
    \caption{\textbf{Drafting depth (MiMo-7B-SFT): $K{=}3, 5, 7$ against the frozen $K{=}1$ head.} Acceptance grows beyond the single-step ceiling, and end-to-end gains peak at $K{=}5$ in both training domains.}
    \label{fig:rq2b}
\end{figure}

\noindent\textbf{RL Training.}
Figure~\ref{fig:rq2b} shows the drafting depth sweep on MiMo-7B-SFT: $\tau$ and the end-to-end step time at $K{=}3$, $5$, and $7$ against the frozen $K{=}1$ head, in both domains.
The $\tau$ trajectories stay separated by depth throughout training, a deeper head drafting more accepted tokens at every step, and none of the three curves has saturated by step 200.
Deeper heads initially cost more per step, but as acceptance grows, the three step-time curves converge far below the frozen baseline: the depth ranking visible on the $\tau$ axis does not carry to the time axis, which determines the preferred depth.

\noindent\textbf{Inference Evaluation.}
Table~\ref{tab:rq2b-eval} reports the held-out evaluation across the four depths, pairing policy quality with $\tau$ on each benchmark.
Every trained depth reaches $\tau$ of at least 3.02, increasing with $K$ on each of the six columns, while policy quality stays on par with the frozen head.
The $\tau$ of $K{=}7$ is the highest in every column, yet the highlighted depth is $K{=}5$: held-out acceptance does not decide the selection, which follows the end-to-end speedups reported in Table~\ref{tab:rq2b-training}, including head-update cost.

\input{tabs/rq2b_eval}

\subsection{Additional Results for Component Ablations (RQ3)}
\label{app:results-rq3}

\begin{figure}[h]
    \centering
    \includegraphics[width=0.7\linewidth]{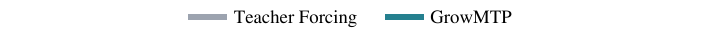}\\[0.3em]
    \begin{subfigure}[c]{0.245\linewidth}
        \centering
        \includegraphics[width=\linewidth]{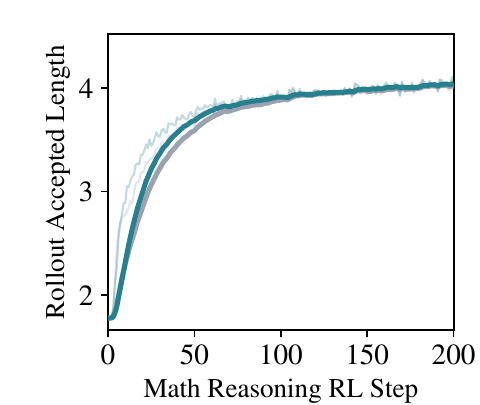}
        \caption{Acceptance length $\tau$.}
        \label{fig:rq3-dpr:tau}
    \end{subfigure}\hfill
    \begin{subfigure}[c]{0.245\linewidth}
        \centering
        \includegraphics[width=\linewidth]{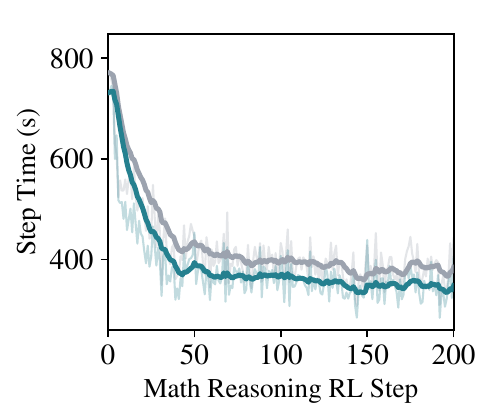}
        \caption{End-to-end step time.}
        \label{fig:rq3-dpr:step}
    \end{subfigure}\hfill
    \begin{subfigure}[c]{0.245\linewidth}
        \centering
        \includegraphics[width=\linewidth]{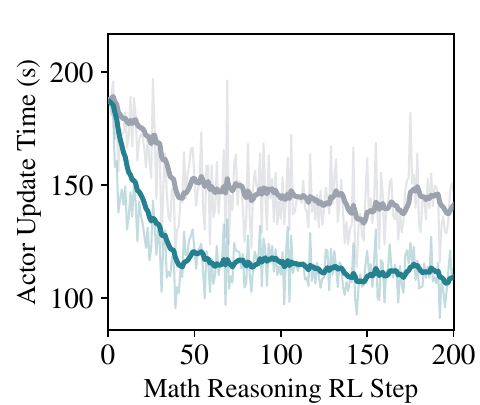}
        \caption{Actor update time.}
        \label{fig:rq3-dpr:update}
    \end{subfigure}\hfill
    \begin{subfigure}[c]{0.245\linewidth}
        \centering
        \includegraphics[width=\linewidth]{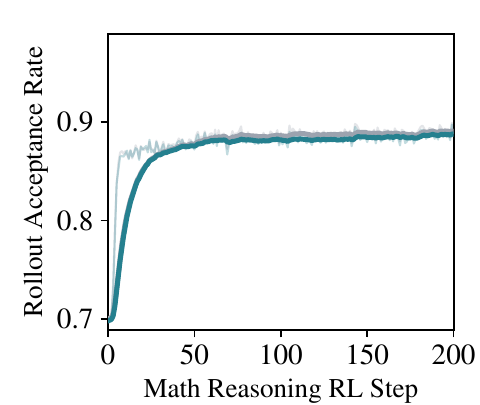}
        \caption{Rollout $\alpha_1$.}
        \label{fig:rq3-dpr:a1}
    \end{subfigure}\\[0.6em]
    \begin{subfigure}[c]{0.245\linewidth}
        \centering
        \includegraphics[width=\linewidth]{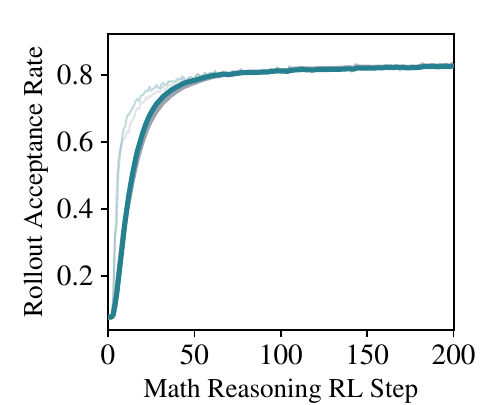}
        \caption{Rollout $\alpha_2$.}
        \label{fig:rq3-dpr:a2}
    \end{subfigure}\hfill
    \begin{subfigure}[c]{0.245\linewidth}
        \centering
        \includegraphics[width=\linewidth]{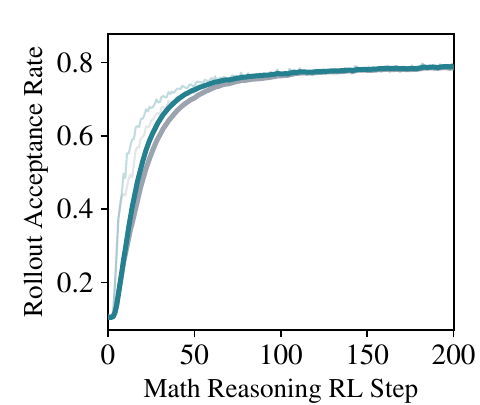}
        \caption{Rollout $\alpha_3$.}
        \label{fig:rq3-dpr:a3}
    \end{subfigure}\hfill
    \begin{subfigure}[c]{0.245\linewidth}
        \centering
        \includegraphics[width=\linewidth]{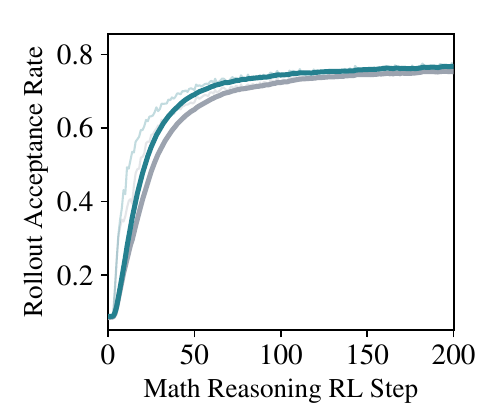}
        \caption{Rollout $\alpha_4$.}
        \label{fig:rq3-dpr:a4}
    \end{subfigure}\hfill
    \begin{subfigure}[c]{0.245\linewidth}
        \centering
        \includegraphics[width=\linewidth]{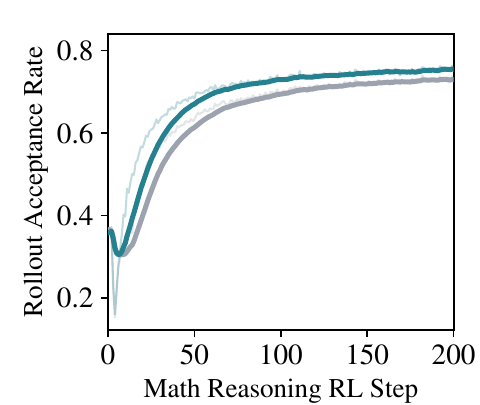}
        \caption{Rollout $\alpha_5$.}
        \label{fig:rq3-dpr:a5}
    \end{subfigure}
    \caption{\textbf{Teacher forcing costs more per step and leaves a residual gap at the deepest draft positions.} MiMo-7B-SFT, $200$ RL steps, $K{=}5$. (a--c) \growmtp reaches the same acceptance length at a lower actor update cost. (d--h) Rollout acceptance $\alpha_k$ separates further with each added depth.}
    \label{fig:rq3-dpr}
\end{figure}

\noindent\textbf{Ablation of Draft-Path Reconstruction.}
Figure~\ref{fig:rq3-dpr} traces both arms: $\tau$, the two timing curves, and rollout acceptance at every draft position.
Teacher forcing pays roughly $30$ extra seconds per actor update, and its bias sits at depth: $\alpha_1$ is indistinguishable between the arms, while the $\alpha_5$ gap narrows without closing.
This is the train--inference inconsistency of \autoref{app:tf} observed in the results.

\noindent\textbf{Ablation of Training Objective.}
Table~\ref{tab:rq3-dca} reports the four objectives at both depths, training and held-out sides.
The $K{=}1$ rows stay within $0.03$ on every $\tau$ column, and at $K{=}5$ DCA holds the highest $\tau$ in every column, extending its training-side lead to all three held-out evaluation benchmarks.

\input{tabs/rq3_dca}

\noindent\textbf{Ablation of Verify-Gated Masking.}
Table~\ref{tab:rq3-vgm} reports the mask switch under both objectives.
Under KL the slowdown is visible already at rollout, while under DCA it falls almost entirely outside rollout, and held-out $\tau$ is never higher with the mask removed on any of the three held-out benchmarks.

\input{tabs/rq3_vgm}

\clearpage
\section{Discussion}
\label{app:discussion}

\subsection{Which Models Can \growmtp Accelerate?}
\label{app:disc-models}

In this section, we discuss the applicability of \growmtp.
The experiments of Section~\ref{sec:experiments} evaluate Qwen3-4B, MiMo-7B-SFT, and Qwen3.5-4B-Base, achieving end-to-end step speedups of $1.60\times$, $1.41\times$, and $1.20\times$ on mathematical reasoning: \growmtp is effective regardless of initial head capability.
For models with pretrained heads, the RL rollout distribution is more concentrated than the broad pretraining data, leaving room for domain-specific adaptation.
Beyond head capability, \growmtp's acceptance-based training objective is model-agnostic: it depends on draft--target token-distribution overlap and the first-rejection index under rejection sampling.
\growmtp can therefore serve as a modular acceleration component in RL training frameworks that employ speculative decoding, particularly for base models without MTP heads, which accelerate without extra pretraining.

\subsection{Does the Acceleration Grow with RL Scale?}
\label{app:disc-scale}

\input{tabs/disc_scale}
The $1.60\times$ of the Qwen3-4B mathematical reasoning run is measured at 500 steps and an 8K rollout length, a small RL scale.
We extrapolate from it along rollout length and training steps: the former increases the proportion of time spent on rollouts, and the latter amortizes the head's growth-phase overhead.
For this projection, we assume that rollout time scales linearly with length while the remaining per-step cost stays constant.
Beyond 500 steps we take the step time measured at the end of training, $398.7$\,s, of which $205.7$\,s is rollout.
We further assume that acceptance length grows neither with rollout length nor with training steps.
The end-to-end speedup then grows monotonically, reaching $1.91\times$ at 16K with 1000 steps and $2.17\times$ at 32K with 2000 steps.
These values are illustrative projections under the stated assumptions. Since the learning curve has not saturated at 500 steps, further head improvement could increase the realized speedup.
Under these assumptions, the ceiling is $2.55\times$, the final measured RL rollout speedup.

\subsection{Does \growmtp Replace Pretrained Draft Heads?}
\label{app:disc-serving}

We freeze the two heads grown from scratch over 500 RL steps, one in each domain, and measure $\tau$ on the held-out benchmarks of both.
Within the training domain, $\tau$ falls from the training rollouts to the held-out benchmarks, 2.91 to 2.32 for the math-trained head and 2.64 to 2.20 for the code-trained head.
Crossing domains is far costlier: the math-trained head falls to 1.45 on code, and the code-trained head to 1.75 on math.
The head receives supervision only from the rollout distribution it must accelerate, and $\tau$ falls as evaluation departs from it.
This specialization is a direct consequence of the narrow-distribution premise: RL acceleration faces a sufficiently concentrated distribution for from-scratch online training to suffice, whereas general serving must cover unforeseen deployment distributions and still relies on pretrained MTP heads with capability learned across broader tasks.

\vspace{\dimexpr\intextsep-5pt\relax}  %
\input{tabs/disc_cross_domain}

\clearpage
\section{\growmtp Implementation Details}
\label{app:impl}

\subsection{Draft-Path Reconstruction}
\label{app:impl-reconstruction}
\noindent\textbf{The Reconstructed Forward.}
Draft-path reconstruction removes the train--inference inconsistency of Section~\ref{sec:method-pipeline} by aligning the training-side computation with inference: the head is trained at the drafted states.
The update phase reconstructs the chain of every recorded cycle: it starts from the backbone hidden recorded at the cycle's start, runs under the current head parameters inside the training graph, and is supervised by the verification signals recorded in the same cycle.
Reconstruction is not redrafting: sampling would not reproduce the rollout's tokens, and with the recursion consuming a token at every position, one changed token changes every hidden state and every cache entry that follows.
The draft tokens therefore turn from outputs into inputs: rollout records the $K$ drafts of every cycle, and they are fed back to the recursion, with nothing sampled.
The hidden states and cache entries along the chain need no recording, since the recursion recomputes them from the same inputs, and every reconstructed state conditions on the recorded prefix of its own draft-then-verify cycle.

\noindent\textbf{The Recorded Supervision.}
The same obstacle appears on the supervision side: $p_{r,k}$ conditions on the drafts themselves (Eq.~\ref{eq:verification-context}), the trajectory retains only the committed path (Eq.~\ref{eq:draft-path-mismatch}), and where the two paths diverge, a target distribution recomputed on the trajectory is no longer the one that verified the head.
The remedy is again recording, not recomputation: verification is itself one target forward over the entire draft path, so it computes the distribution at every draft position, and rollout stores these beside the drafts, one per position.
The two remedies interlock in the conditioning: the reconstructed $q_{r,k}$ of Eq.~\ref{eq:draft-path-reconstruction} and the recorded $p_{r,k}$ condition on the same recorded prefix $[\mathbf{c}_r, d_{r,1},\ldots,d_{r,k-1}]$, and the pairing Section~\ref{sec:method-pipeline} defines at the token level becomes $K$ concrete pairs per cycle in the training graph, consumed directly by the loss of Section~\ref{sec:method-dca}.
The reconstruction is self-contained: a chain reads only its own cycle's records and the shared prefix, and no chain uses another chain's states.

\vspace{-\baselineskip}
\enlargethispage{\baselineskip}
\subsection{Implementing Draft-Path Reconstruction in verl}
\label{app:impl-recording}
\noindent\textbf{Recording in Rollout.}
The record of a cycle holds five items: the backbone hidden at the cycle's start, the $K$ drafts, the verification distributions, the accepted length, and the absolute position of the start.
The hidden is captured rather than recomputed: it is the very tensor the engine's drafting consumed, and the reconstruction therefore starts from a state numerically identical to drafting's.
The verification distributions are stored as the target's top-64 log-probabilities in bf16: keeping the full distribution at every draft position would scale with the vocabulary, and the truncation fixes the record's width at 64.
The acceptance decisions compress to one integer per cycle: the accepted positions form a prefix, so a single length, the further token counted, restores the decision at every position.
The record carries the absolute position of the start, and the reconstruction puts each chain back where it sat in the sequence.
Each sequence additionally stores the rollout-time target hidden states and token ids for the prompt and all committed tokens.
These recorded inputs are used to prefill the MTP head and rebuild the shared prefix KV cache attended by each reconstructed draft chain.

The stored top-64 log-probabilities retain full-vocabulary normalization rather than being renormalized over the retained set.
During head training, the draft distribution is normalized over the full vocabulary and gathered at the target's top-64 indices; the remaining mass of each distribution forms a residual bin.
The resulting overlap upper-bounds the full-vocabulary overlap by at most the target mass outside the top-64, so the error is small when the retained target mass is high.
We quantify its effect on three checkpoints from Qwen3-4B Math RL: random initialization, step 100, and step 500.

Against the full-vocabulary reference, mean absolute overlap errors range from $1.8\times10^{-6}$ to $5.7\times10^{-6}$, and the mean per-batch cosine similarity between sparse and full-vocabulary head gradients is at least $0.99996$ at each checkpoint.
Thus, top-64 sparsification closely preserves overlap values and gradient directions.
Rollout verification still uses the full distribution, so the approximation affects only the draft-head training objective without affecting rollout-time verification decisions.

\noindent\textbf{Integration into the RL Loop.}
The head is a submodule of the FSDP-wrapped actor model, so the shared optimizer of Appendix~\ref{app:setup-head} is structure rather than wiring.
Deployment rides the same structure: the weight synchronization that carries the updated actor into the drafting engine carries the head with it, and after every step the engine drafts with the head just trained.
The recording sits on the same loop: signals are captured inside the engine's verification and delivered to the update phase with the rollout batch.
The update phase therefore receives one record per cycle: a cycle commits $\tau$ tokens on average, so the records are fewer than the response's token positions by a factor of $\tau$.

\clearpage
\subsection{Parallelizing \growmtp Training}
\label{app:impl-parallel}
\begin{figure}[h]
    \centering
    \includegraphics[width=0.82\linewidth]{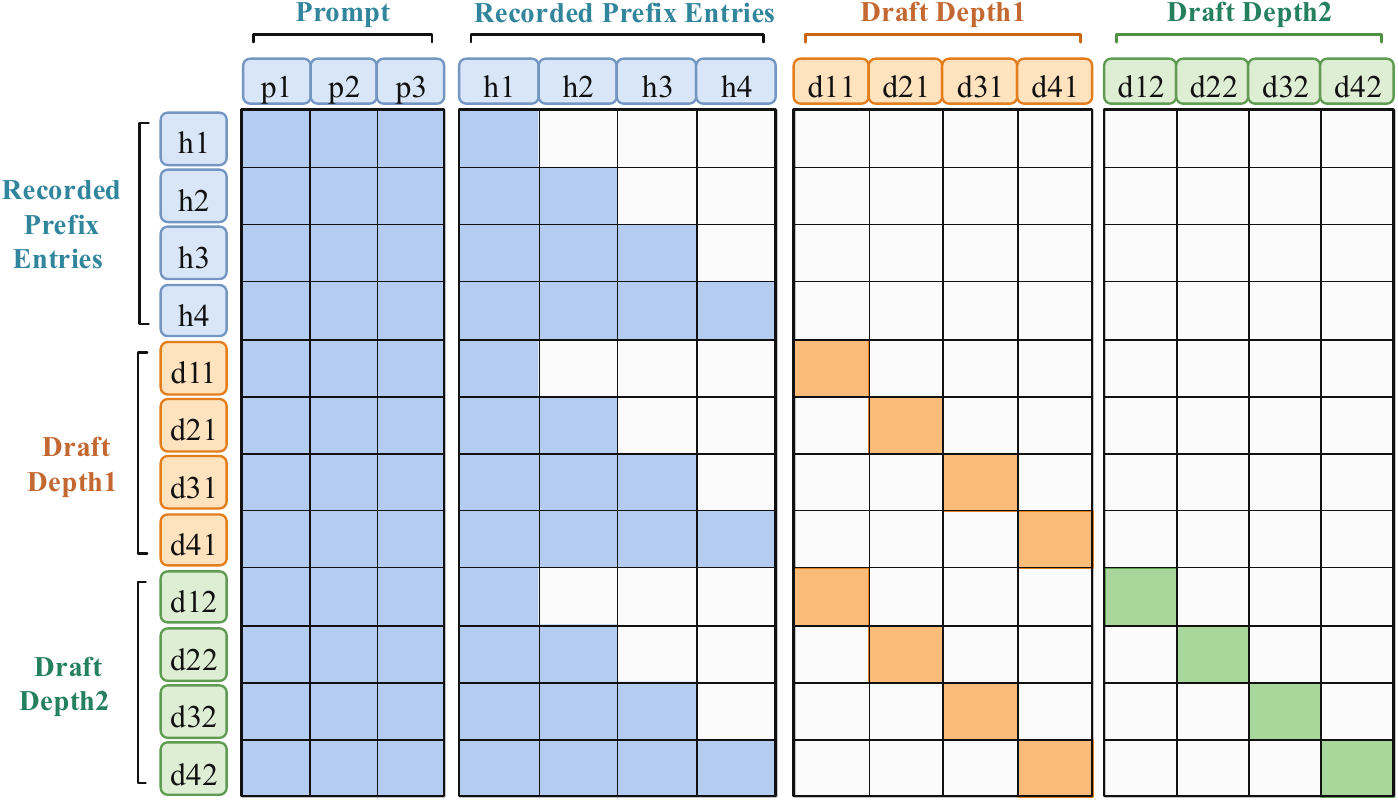}
    \vspace{-6pt}
    \caption{\textbf{The attention mask of one batched reconstruction forward.} Every chain attends over one shared cache: causal over the full committed-token prefix, diagonal over the draft entries of every depth. Each query sees exactly the context of chain-by-chain execution on recorded cycle inputs.}
    \label{fig:attn-mask}
    \vspace{-0.4cm}
\end{figure}
\noindent\textbf{Serial Depth, Parallel Cycles.}
Executed chain by chain, the reconstruction of Appendix~\ref{app:impl-reconstruction} costs one forward per draft: for a response of three thousand cycles at $K=5$, that is fifteen thousand single-token forwards in sequence.
Each of these forwards is indexed by the pair $(r,k)$, with the cycle index $r$ locating the chain within the response and the depth $k$---the draft position within its cycle---locating the forward within the chain.
Along $k$ there is no parallelism to extract: Eq.~\ref{eq:draft-path-reconstruction} takes the state at depth $k-1$ as input at depth $k$, and the $K$ forwards of one chain can only run in order.
Along $r$ no such dependence exists: chains share only recorded inputs, never each other's computed states, so the forwards of every cycle at a fixed depth admit a single batched execution.
Serialization is therefore confined to depth: any schedule spends at least $K$ forwards one after another, and the width available at each of them is the number of cycles in the response, each advanced independently.

\noindent\textbf{Parallelism across Cycles.}
The reconstruction therefore runs as $K$ batched forwards: the forward at depth $k$ advances all chains by one draft, and the serial count of the response falls from fifteen thousand to $K$.
In one batched forward, visibility no longer follows from packing order: each query must be restricted to the context chain-by-chain execution would give it, a restriction the attention mask of Figure~\ref{fig:attn-mask} carries.
All chains attend over one shared cache, with causal visibility over the full committed-token prefix and diagonal visibility over the draft entries, so no chain sees another's drafts.
For each cycle, the prefilled cache is restricted to its committed prefix, while the cycle's own draft KV entries are recomputed by feeding back its recorded draft tokens in order.
The mask is additive and needs no custom attention kernel: standard scaled dot-product attention consumes it directly.
Positions are taken from the record rather than from the packing: every entry enters the rotary encoding at the position it held at rollout, the depth-$k$ draft of cycle $r$ sitting $k$ places past its cycle's start.
Merging changes the execution schedule while preserving each chain's computation.

\noindent\textbf{Bounding the Memory.}
What merging does change is retention: a forward's activations stay resident until its backward runs, so cycles that advance together are held in memory together.
At full width the resident state is the whole response---every chain at every depth, held for one backward---and at training lengths it exceeds the accelerator's memory.
Bounding memory is bounding the span of one backward: the reconstruction proceeds in chunks of 1024 consecutive cycles, each chunk running its $K$ forwards and its own backward and releasing its activations before the next begins.
Chunking parallelizes nothing: it only sets how many cycles are resident at once.
Across chunks the entries of the recorded hiddens accumulate, entering later chunks as constants, while a chunk's draft entries---entries the diagonal pattern shows to no later chain---are dropped at its end.
Each chunk's loss is weighted by its share of the response's positions, and the gradients accumulated across chunks equal those of one backward over the response's loss: cross-chunk entries are constants, the chunks' graphs are disjoint, and gradients over disjoint graphs add.
A response of three thousand cycles spans three chunks: fifteen batched forwards, where chain-by-chain reconstruction ran fifteen thousand.

\clearpage
\section{\growmtp Training Algorithm}
\label{app:algo}
Algorithm~\ref{alg:scalemtp} states one \growmtp training step: against the two-phase step of Section~\ref{sec:prelim-rl}, the additions are the records $\{R_r\}$ of line~\ref{algline:specdec} and the head update of lines~\ref{algline:for}--\ref{algline:head}, and no other line computes anything new.
The record $R_r$ of cycle $r$ collects the five items of Appendix~\ref{app:impl-recording}; the update consumes four of them by name---the backbone hidden $b_r$ at the cycle's start, the drafts $\mathbf{d}_r$, the verification distributions $p_{r,1:K}$, and the first rejection $j_r$, read from the accepted length---while the start position is consumed inside the reconstruction alone.
The inner loop runs the recursion of Eq.~\ref{eq:draft-path-reconstruction} on recorded inputs: the chain starts from the hidden $b_r$, which enters the graph detached, and the recorded start position gives the chain its place in the sequence.
\textsc{SpecDecode} abbreviates the draft-then-verify generation of Section~\ref{sec:prelim-specdec}, inside which the records are captured cycle by cycle.
\textsc{Sync} abbreviates the weight synchronization of Appendix~\ref{app:impl-recording}, and all remaining symbols are those of Sections~\ref{sec:prelim} and~\ref{sec:method}.

\begin{algorithm}[h]
\caption{One \growmtp training step.}
\label{alg:scalemtp}
\algrenewcommand{\algorithmiccomment}[1]{\hfill{\color{green!50!black}\hypersetup{linkcolor=green!50!black}// #1}}
\begin{algorithmic}[1]
\Require policy $\pi_\theta$, draft head $h_\phi$, a batch of prompts $X$
\Ensure $\theta$ and $\phi$ updated and synchronized into the drafting engine
\Statex {\color{green!50!black}// \textit{Rollout (drafting engine)}}
\State $Y,\,\{R_r\} \leftarrow{}$\textsc{SpecDecode}$(X;\,\pi_\theta,\,h_\phi)$ \Comment{Sec.~\ref{sec:prelim-specdec}; App.~\ref{app:impl-recording}}\label{algline:specdec}
\State deliver $Y$, its rewards, and $\{R_r\}$ to the update phase
\Statex {\color{green!50!black}// \textit{Update (actor)}}
\State update $\pi_\theta$ by the RL objective on $Y$ \Comment{Sec.~\ref{sec:prelim-rl}}
\For{each recorded cycle $r$}\label{algline:for} \Comment{batched across cycles, App.~\ref{app:impl-parallel}}
    \For{$k = 1, \ldots, K$} \Comment{Eq.~\ref{eq:draft-path-reconstruction}; serial in $k$}
        \State $q_{r,k} \leftarrow h_\phi(\cdot \mid s_{r,k})$
        \State $s_{r,k+1} \leftarrow [s_{r,k},\, d_{r,k}]$
    \EndFor
    \State $\alpha_{r,k} \leftarrow 1-\mathrm{TV}(p_{r,k},\,q_{r,k})$, $k=1,\ldots,K$ \Comment{Eq.~\ref{eq:alpha}}
    \State $\mathcal{L}_r \leftarrow \mathcal{L}_{\mathrm{DCA}}^{\mathrm{VGM}}(\alpha_{r,1:K};\,j_r)$ \Comment{Eq.~\ref{eq:vgm}: chain sum truncated at $\min(j_r,K)$}
\EndFor
\State update $h_\phi$ by $\nabla_\phi$ of the mean of $\mathcal{L}_r$ over the batch's cycles \Comment{the gradient reaches only $\phi$}\label{algline:head}
\State \textsc{Sync}$(\theta,\,\phi)$ \Comment{line~\ref{algline:specdec} of the next step drafts with this $\phi$}
\end{algorithmic}
\end{algorithm}

\clearpage
\section{Train--Inference Inconsistency under Teacher Forcing}
\label{app:tf}

This appendix identifies the cycle-boundary and KV-cache mismatch between committed-response teacher forcing and rollout-time drafting, and traces its dependence on draft depth.
The cause is one mechanism: inference restarts the head's recursion, chain and cache alike, only at cycle boundaries, while teacher forcing, consuming the flat response alone, makes every position a start and restarts nowhere.
Draft-path reconstruction resolves the inconsistency by recomputing exactly the states inference runs, and no teacher-forcing mask can replace the reconstruction of the drafting states.

\noindent\textbf{Training Under Teacher Forcing.}
Teacher forcing consumes only the final response $y_1,\ldots,y_T$, and none of the cycle structure recorded during the rollout.
The head is a single module applied in a loop, consuming at depth $k$ and position $i$ the depth-$(k{-}1)$ output and the token embedding at the position before it, and attending the key-value entries held in the cache at earlier positions of the response,
\begin{equation}
    g_i^{(k)} \;=\; f_\phi\bigl(g_{i-1}^{(k-1)},\, e(y_{i-1}) \,;\, \mathcal{K}_i^{(k)}\bigr),
    \qquad k = 1, \ldots, K,
    \label{eq:tf-loop}
\end{equation}
where $e$ maps a token to its embedding, $g^{(0)}$ are the hidden states the target model produced during the rollout forward, and $g_i^{(k)}$ plays the role the sequences $s_{r,k}$ play in Eq.~\ref{eq:draft-path-reconstruction}.
The cache grows by one entry per position, written by the head's key-value projection $c_\phi$ from the same pair of head inputs,
\begin{equation}
    \mathcal{K}_i^{(k)} \;=\; \mathcal{K}_{i-1}^{(k)} \,\cup\, \bigl\{ c_\phi\bigl(g_{i-1}^{(k-1)},\, e(y_{i-1})\bigr) \bigr\},
    \label{eq:tf-kv}
\end{equation}
starting from the entries of the prompt in $\mathcal{K}_0^{(k)}$: the cache attended at depth $k$ is thus made entirely of entries computed from depth-$(k{-}1)$ features.
The $K$ causal forwards yield at most $T\cdot K$ training pairs $\bigl(g_i^{(k)}, y_i\bigr)$, whose unrolled recursion traces each trained state back to the target's hidden features,
\begin{equation}
    g_{i-k}^{(0)} \;\xrightarrow{\;f_\phi\;}\; g_{i-k+1}^{(1)} \;\xrightarrow{\;f_\phi\;}\; \cdots \;\xrightarrow{\;f_\phi\;}\; g_{i-1}^{(k-1)} \;\xrightarrow{\;f_\phi\;}\; g_i^{(k)}.
    \label{eq:tf-chain}
\end{equation}
The chain starts at position $i-k$, from the target's own feature there: teacher forcing makes every position the start of a chain, and sets the start by depth alone rather than by recorded cycle boundaries.

\noindent\textbf{Inference Under Speculative Decoding.}
At inference the head runs the same recursion and restarts it at one kind of position only: the end of the committed prefix, once per cycle.
Cycle $r$ begins drafting at context length $m_r$, and every position behind this boundary carries a committed token together with its target feature $g^{(0)}$.
To draft position $m_r{+}k$, the head applies $f_\phi$ $k$ times, and the first application consumes the target feature at the boundary before it uses its own recursive features,
\begin{equation}
\begin{aligned}
    \bar{g}_{r,1} &= f_\phi\bigl(g_{m_r}^{(0)},\, e(y_{m_r}) \,;\, \bar{\mathcal{K}}_{r,1}\bigr), \\
    \bar{g}_{r,k} &= f_\phi\bigl(\bar{g}_{r,k-1},\, e(d_{r,k-1}) \,;\, \bar{\mathcal{K}}_{r,k}\bigr),
    \qquad k = 2, \ldots, K.
\end{aligned}
\label{eq:tf-restart}
\end{equation}
The first line shows where the recursion begins again: $\bar{g}_{r,k}$, the state at which the head evaluates $h_\phi(\cdot\mid s_{r,k})$ in Eq.~\ref{eq:draft-path-reconstruction}, is a chain of length $k$ starting at $m_r$, and the only chain the cycle runs.
The tokens this chain consumes are the head's own drafts, which follow the committed response up to the first rejection, the divergence Eq.~\ref{eq:draft-path-mismatch} records.
The restart resets the cache the recursion has written along with the chain's start: cycle $r$ closes by committing up to the next boundary $m_{r+1}$, and the cache the last draft state read stands above the cache the next cycle's first drafting application reads,
\begin{equation}
\begin{aligned}
    \bar{\mathcal{K}}_{r,K} &= \bigl\{\, \ldots,\; c_\phi\bigl(g_{m_r}^{(0)},\, e(y_{m_r})\bigr),\; c_\phi\bigl(\bar{g}_{r,1},\, e(d_{r,1})\bigr),\; \ldots,\; c_\phi\bigl(\bar{g}_{r,K-1},\, e(d_{r,K-1})\bigr) \,\bigr\}, \\
    \bar{\mathcal{K}}_{r+1,1} &= \bigl\{\, \ldots,\; c_\phi\bigl(g_{m_r}^{(0)},\, e(y_{m_r})\bigr),\; c_\phi\bigl(g_{m_r+1}^{(0)},\, e(y_{m_r+1})\bigr),\; \ldots,\; c_\phi\bigl(g_{m_{r+1}}^{(0)},\, e(y_{m_{r+1}})\bigr) \,\bigr\}.
\end{aligned}
\label{eq:tf-restart-kv}
\end{equation}
Over the accepted range the two rows hold the same tokens, the committed $y_{m_r+l}$ being the draft $d_{r,l}$, so the restart replaces the feature alone: $\bar{g}_{r,l}$ becomes $g_{m_r+l}^{(0)}$.
This composition then holds at every depth of every cycle: the entries behind the boundary are computed from the target's features, and the head's recursive features enter the cache only through the current chain's writes.
The restart does not depend on rejection: a rollout that accepts every draft still closes its cycle after $K{+}1$ committed tokens, with $m_{r+1} = m_r{+}K{+}1$, and recursion restarts at the boundary to draft the next cycle.

\clearpage
\noindent\textbf{Train--Inference Inconsistency.}
Take a position $i$ that cycle $r$ both drafted and committed: teacher forcing trains $K$ states there, starts $i{-}1$ through $i{-}K$, while inference computes exactly one, reached at depth $k = i{-}m_r$ by the only chain the cycle runs from its start $m_r$.
Inference never computes the states at the other starts: a mid-cycle position is never a restart, and the chain from an earlier boundary leaves the committed response before reaching $i$ (Eq.~\ref{eq:draft-path-mismatch}).
The remaining pair shares start and tokens alike, the drafts along the drafted chain being committed tokens here, so the two computation graphs can differ only through their caches.
A single boundary entry already separates the two caches:
\begin{equation}
\begin{aligned}
    &\text{teacher forcing training:} && c_\phi\bigl(g_{m_r}^{(k-1)},\, e(y_{m_r})\bigr) \;\in\; \mathcal{K}_i^{(k)}
      && \text{(Eq.~\ref{eq:tf-kv})}, \\
    &\text{speculative decoding inference:} && c_\phi\bigl(g_{m_r}^{(0)},\, e(y_{m_r})\bigr) \;\in\; \bar{\mathcal{K}}_{r,k}
      && \text{(Eq.~\ref{eq:tf-restart-kv})}.
\end{aligned}
\label{eq:tf-gap}
\end{equation}
The same projection of the same token, the difference contracted to a single superscript, $k{-}1$ against $0$: the entries agree only at $k = 1$, where both caches are computed from the target's features end to end, and the matching states across the whole response occur only at each cycle's first draft position:
\begin{equation}
    g_i^{(k)} \;=\; \bar{g}_{r,\,i-m_r}
    \qquad\Longleftrightarrow\qquad
    k \;=\; i-m_r \;=\; 1.
    \label{eq:tf-coincide}
\end{equation}
That is one state per cycle, out of the $T\cdot K$ states teacher forcing trains, and the rest update the same $\phi$, teaching it computations inference never runs.
At depth one every trained state still has the drafted form, one application of $f_\phi$ to a target feature under a cache computed from the target's features: the two schemes differ only in which starts the rollout realizes.
At depth $k$ the trained chain spans $k$ positions and crosses a cycle boundary whenever one falls among them, carrying head-computed features through a position at which inference restarts, and a longer chain straddles a boundary more often: the departure from the states inference runs grows along the depth axis.
And this is the very axis a deeper draft must scale: the deeper the head drafts, the further from its inference states teacher forcing trains it, so deeper drafting increases the mismatch between training and inference states.

\noindent\textbf{\growmtp Addresses the Inconsistency.}
\growmtp trains the head at the states of Eq.~\ref{eq:tf-restart} themselves: the update phase restarts the recursion of Eq.~\ref{eq:draft-path-reconstruction} at each recorded boundary (Algorithm~\ref{alg:scalemtp}, App.~\ref{app:impl-reconstruction}), feeding the recorded drafts back in order, and its batched execution changes the schedule of this computation and nothing that is computed (App.~\ref{app:impl-parallel}).
Every separation drawn above closes: the chain starts only at the recorded boundary feature $g_{m_r}^{(0)}$, not at every position, the tokens it consumes are the recorded drafts themselves, and the cache rebuilt behind the boundary holds entries computed from the target's features, the head's recursion entering only through the chain's own writes---precisely the composition of the inference cache.
In Eq.~\ref{eq:tf-gap} the training-side entry now carries superscript $0$, the very entry inference attends, and where Eq.~\ref{eq:tf-coincide} left teacher forcing one coincidence per cycle, all $K$ states of the cycle coincide.
No mask over the teacher-forcing forward substitutes for this: a mask selects within what teacher forcing supplies, yet the forward, restarting nowhere, computes no drafted state to keep, and the flat response it consumes carries no cycle for the gate of Eq.~\ref{eq:vgm} to truncate---what produces the drafted states is reconstruction itself.
The inconsistency this appendix constructs therefore does not arise under \growmtp: training uses inference states at all depths.

\clearpage
\section{One-Step Lag Bound under Rejection Sampling}
\label{app:lag}

The training loop of Section~\ref{sec:method-pipeline} retains a single asymmetry between successive RL steps: the head that drafts at step $t{+}1$ was trained on verification signals recorded at step $t$, one policy update earlier. The head is therefore always trained against a target one update older than the one it drafts for, and this \emph{one-step lag} may degrade the acceptance rate that training establishes. This appendix characterizes the sensitivity of the acceptance chain to this lag through the target shift in a single policy update.

\noindent\textbf{Setup.} Fix policy update $t \to t{+}1$ and conditioning state $s$, and consider these three distributions:
\begin{equation}
    \scalebox{0.92}{$\displaystyle
    p^{(t)}(\cdot\mid s) = \operatorname{softmax}\bigl(z_{\theta_t}(s)\bigr),
    \quad
    p^{(t+1)}(\cdot\mid s) = \operatorname{softmax}\bigl(z_{\theta_{t+1}}(s)\bigr),
    \quad
    q(\cdot\mid s) = h_{\phi_{t+1}}(\cdot\mid s):
    $}
    \label{eq:lag-dists}
\end{equation}
the old target, from which the head's training signals were recorded, the new target, against which the head's drafts are verified at step $t{+}1$, and the head distribution, which generates those drafts. Eq.~\ref{eq:alpha} assigns the head two acceptance rates at the same conditioning state, one for each of the two targets:
\begin{equation}
    \tilde{\alpha}(s) = 1 - \mathrm{TV}\bigl(p^{(t)}(\cdot\mid s),\, q(\cdot\mid s)\bigr),
    \qquad
    \alpha(s) = 1 - \mathrm{TV}\bigl(p^{(t+1)}(\cdot\mid s),\, q(\cdot\mid s)\bigr),
    \label{eq:lag-rates}
\end{equation}
the training-time rate $\tilde{\alpha}$, taken against the target the head was trained on, and the deployment rate $\alpha$, taken against the target that verifies its drafts. The cost of the lag is the difference between the two.

\noindent\textbf{Single-Position Bound.} By Eq.~\ref{eq:lag-rates}, the cost is a difference of two distances from $q$, and the triangle inequality bounds it from above by using the old target distribution as the intermediate distribution:
\begin{equation}
    \scalebox{0.86}{$\displaystyle
    \tilde{\alpha}(s) - \alpha(s) = \mathrm{TV}\bigl(p^{(t+1)}(\cdot\mid s),\, q(\cdot\mid s)\bigr) - \mathrm{TV}\bigl(p^{(t)}(\cdot\mid s),\, q(\cdot\mid s)\bigr) \le \mathrm{TV}\bigl(p^{(t+1)}(\cdot\mid s),\, p^{(t)}(\cdot\mid s)\bigr),
    $}
    \label{eq:lag-pos}
\end{equation}
which holds for every state $s$ and uniformly in the head distribution $q$: the deployment rate falls below the training-time rate by at most the single-update displacement of the target. Since $\mathrm{TV}$ is a metric, the same argument with the roles of $p^{(t)}$ and $p^{(t+1)}$ exchanged bounds $\alpha(s) - \tilde{\alpha}(s)$ by the same quantity, so the bound controls both increases and decreases in acceptance at the same fixed state. 

\noindent\textbf{Per-Cycle Bound.} For the per-cycle acceptance-chain value defined in Eq.~\ref{eq:chain}, define the target shift as
\begin{equation}
    \bar{\varepsilon}_t \;\triangleq\; \max_{k \le K}\; \mathbb{E}_{s_k}\Bigl[\mathrm{TV}\bigl(p^{(t+1)}(\cdot\mid s_k),\, p^{(t)}(\cdot\mid s_k)\bigr)\Bigr] \;=\; \frac{1}{2}\,\max_{k \le K}\; \mathbb{E}_{s_k}\,\mathbb{E}_{y \sim p^{(t)}}\bigl|\,\rho(y) - 1\,\bigr|,
    \label{eq:lag-shift}
\end{equation}
where $\mathbb{E}_{s_k}$ averages over the states conditioning draft position $k$ in the cycles of the step-$t{+}1$ rollout and $\rho(y) = p^{(t+1)}(y\mid s_k)\,/\,p^{(t)}(y\mid s_k)$ is the token importance ratio between successive policies.
\begin{equation}
    \mathrm{TV}(p, q) \;=\; \frac{1}{2}\sum_y p(y)\,\biggl|\,\frac{q(y)}{p(y)} - 1\,\biggr|,
    \label{eq:lag-ratio}
\end{equation}
valid here since softmax distributions have full support. The first form is used in the derivation, and the second expresses $\bar{\varepsilon}_t$ through the importance ratio and underlies its estimation below. Fix a cycle of the step-$t{+}1$ rollout with conditioning states $s_1, \ldots, s_K$, and write $\alpha_i = \alpha(s_i)$, $\tilde{\alpha}_i = \tilde{\alpha}(s_i)$, and $\delta_i = \mathrm{TV}\bigl(p^{(t+1)}(\cdot\mid s_i),\, p^{(t)}(\cdot\mid s_i)\bigr)$. Since every factor lies in $[0,1]$ and $\tilde{\alpha}_n - \alpha_n \le \delta_n$ by Eq.~\ref{eq:lag-pos}, the difference between products can be expanded as the following telescoping sum for each $l \le K$:
\begin{equation}
    \prod_{i=1}^{l} \tilde{\alpha}_i \;-\; \prod_{i=1}^{l} \alpha_i \;=\; \sum_{n=1}^{l} \Bigl(\prod_{i<n} \tilde{\alpha}_i\Bigr)\bigl(\tilde{\alpha}_n - \alpha_n\bigr)\Bigl(\prod_{n<i\le l} \alpha_i\Bigr) \;\le\; \sum_{n=1}^{l} \delta_n.
    \label{eq:lag-telescope}
\end{equation}
Summing over $l = 1, \ldots, K$, taking expectations over the cycles of the rollout, and bounding each $\mathbb{E}[\delta_n]$ by $\bar{\varepsilon}_t$, the shifts are counted $\sum_{l=1}^{K} l = K(K+1)/2$ times across the chain terms, yielding
\begin{equation}
    \mathbb{E}\Biggl[\,\sum_{l=1}^{K} \prod_{i=1}^{l} \tilde{\alpha}_i\Biggr] \;-\; \mathbb{E}\Biggl[\,\sum_{l=1}^{K} \prod_{i=1}^{l} \alpha_i\Biggr] \;\le\; \frac{K(K+1)}{2}\,\bar{\varepsilon}_t.
    \label{eq:lag-chain}
\end{equation}
The second expectation is the expected acceptance-chain value of Eq.~\ref{eq:chain} at step $t{+}1$, and the first is the same quantity measured against the training-time target: it is the acceptance-chain quantity underlying the DCA loss, with both targets evaluated at the same draft states from the new rollout.

\clearpage

\noindent\textbf{Magnitude of $\bar{\varepsilon}_t$.} The shift is a property of the RL update alone, and its leading-order magnitude follows from two classical facts. First, by Eq.~\ref{eq:lag-dists} the two targets are softmax outputs of the same network at neighboring parameter points, so their KL divergence admits the following second-order expansion in the parameter update $\Delta\theta = \theta_{t+1} - \theta_t$, evaluated at the same conditioning state $s$:
\begin{equation}
    \mathrm{KL}\Bigl(p^{(t)}(\cdot\mid s)\,\Big\|\, p^{(t+1)}(\cdot\mid s)\Bigr) \;=\; \frac{1}{2}\,\Delta\theta^{\top} F(s)\, \Delta\theta \;+\; O\bigl(\|\Delta\theta\|^{3}\bigr),
    \label{eq:lag-kl}
\end{equation}
where $F(s)$ is the Fisher information matrix of the target model at state $s$, the Hessian of the divergence at zero displacement. Second, Pinsker's inequality $\mathrm{TV} \le \sqrt{\mathrm{KL}/2}$, Jensen's inequality, and the maximization over draft positions in Eq.~\ref{eq:lag-shift} convert Eq.~\ref{eq:lag-kl} into this bound on target shift:
\begin{equation}
    \bar{\varepsilon}_t \;\le\; \frac{1}{2}\sqrt{\Delta\theta^{\top} \bar{F}\, \Delta\theta \;+\; O\bigl(\|\Delta\theta\|^{3}\bigr)},
    \label{eq:lag-fisher}
\end{equation}
where $\bar{F}$ is the Fisher matrix averaged over the conditioning states at the maximizing position: to leading order, the shift is at most half the length of the update measured in the Fisher metric, and vanishes linearly with the learning rate. The clipped surrogate, small learning rate, and gradient-norm clipping aim to limit policy movement but do not determine $\bar{\varepsilon}_t$, the per-update target-policy shift.

\noindent\textbf{Measuring $\bar{\varepsilon}_t$.} The ratio form of Eq.~\ref{eq:lag-shift} makes the shift directly measurable: rollout tokens are sampled from $p^{(t)}$, so the sample mean of $\tfrac{1}{2}|\rho - 1|$ over a training batch estimates the inner expectation unbiasedly, at the cost of one additional scoring pass of the batch under the updated policy. In our runs, the measured per-token $\mathrm{KL}\bigl(p^{(t)} \,\|\, p_{\mathrm{ref}}\bigr)$ remains below $0.007$ nats throughout training. This fixed-reference KL provides contextual evidence that the policy trajectory remains localized, although it does not directly estimate the consecutive-policy shift $\bar{\varepsilon}_t$ in the acceptance-chain bound of Eq.~\ref{eq:lag-chain}.

\noindent\textbf{Conclusion.} Eq.~\ref{eq:lag-chain} shows that the acceptance-chain shift is controlled by the consecutive-policy shift $\bar{\varepsilon}_t$, with a worst-case bound of $\tfrac{K(K+1)}{2}\bar{\varepsilon}_t$. Because the head is refreshed after every RL step, the bound depends only on the current policy update rather than accumulated training drift; the sustained acceptance gains in our runs show that this bounded staleness does not prevent effective online adaptation. Under rejection sampling, Eq.~\ref{eq:alpha} makes the acceptance rate $1$-Lipschitz in the target distribution under total variation, with both the draft distribution and the conditioning state held fixed.

\clearpage
\section{Optimization Analysis of Training Objective}
\label{app:dca}

The most direct way to improve rollout efficiency is to train the MTP head with a differentiable surrogate for the acceptance length, the TV-chain objective $\mathcal{L}_{\mathrm{TV}}$ of Section~\ref{sec:method-dca}, which prior work adopts.
This surrogate is multiplicative in the per-position acceptance probabilities, so its value on a draft step measures how well the head already drafts there, and a direct design carries that value into the gradient.
This appendix compares TV and DCA gradients (Eq.~\ref{eq:dca}) for individual draft steps and batches under SGD and Adam, characterizing how the logarithm changes relative cycle weights.

\noindent\textbf{Setup.} The head enters either objective only through the acceptance probabilities of one draft-then-verify cycle, which compound into the per-cycle acceptance-chain surrogate specified in Eq.~\ref{eq:chain},
\begin{equation}
    M \;=\; \sum_{l=1}^{K} \prod_{i=1}^{l} \alpha_i,
    \qquad
    \mathcal{L}_{\mathrm{TV}} \;\triangleq\; -M,
    \qquad
    \mathcal{L}_{\mathrm{DCA}} \;\triangleq\; -\log M,
    \label{eq:dca-objectives}
\end{equation}
where $\mathcal{L}_{\mathrm{TV}}$ differs by an affine constant from the normalized form $1 - M/K$ in which prior work writes it.
Since softmax distributions have full support, $\alpha_k \in (0,1]$ and $M \in (0,K]$, the lower end approached by a head whose drafts are rejected at the first position and the upper end attained by one whose drafts are always accepted.
The verify-gated form of Eq.~\ref{eq:vgm} replaces $K$ by $\min(j,K)$ in the chain sum, which changes the value of $M$ and the upper end of this range and leaves the statements below unaffected.
Both objectives are built from the same acceptance probabilities of the same cycle, so the logarithm of the shared chain value accounts for every difference between the two objectives.

\noindent\textbf{Gradient Analysis.} To compare the gradients of the two objectives, differentiate each of them with respect to the individual acceptance probabilities.
The recorded target distributions enter both objectives as constants, and the gradients below are taken with respect to the head parameters $\phi$ alone.
The derivative of $M$ at a single position follows from its multilinearity in the acceptance probabilities: $\alpha_k$ occurs once in every term whose index is at least $k$ and in no other term, and factoring it out separates the product over all preceding positions from the sum of products over subsequent positions:
\begin{equation}
    \frac{\partial M}{\partial \alpha_k}
    \;=\; \sum_{l=k}^{K} \prod_{i \le l,\, i \ne k} \alpha_i
    \;=\; \Bigl(\prod_{i<k} \alpha_i\Bigr) \sum_{l=k}^{K} \prod_{i=k+1}^{l} \alpha_i.
    \label{eq:dca-partial}
\end{equation}
The remaining sum is the chain quantity of Eq.~\ref{eq:dca-objectives} formed over the positions after $k$, a sum of $K-k+1$ terms, with the first term equal to one and each remaining term between zero and one:
\begin{equation}
    D_k \;\triangleq\; \sum_{l=k}^{K} \prod_{i=k+1}^{l} \alpha_i \;\in\; [1,\, K-k+1].
    \label{eq:dca-tail}
\end{equation}
Since $\mathcal{L}_{\mathrm{TV}} = -M$ and $\mathcal{L}_{\mathrm{DCA}} = -\log M$, the coefficients the two objectives place on position $k$ are $\partial M / \partial \alpha_k$ and $M^{-1}\, \partial M / \partial \alpha_k$, respectively. The derivative of $M$ yields the following coefficients:
\begin{equation}
    -\frac{\partial \mathcal{L}_{\mathrm{TV}}}{\partial \alpha_k}
    \;=\; \Bigl(\prod_{i<k} \alpha_i\Bigr) D_k,
    \qquad
    -\frac{\partial \mathcal{L}_{\mathrm{DCA}}}{\partial \alpha_k}
    \;=\; \Bigl(\prod_{i<k} \alpha_i\Bigr) \frac{D_k}{M},
    \label{eq:dca-coeff}
\end{equation}
where $D_k/M$ matches the coefficient of Eq.~\ref{eq:weight}, its denominator shared across positions and its numerator at most $K-k+1$, which bounds its variation across positions by a factor of $K$.
Assembling the positions into the gradient with respect to $\phi$ places the common positive factor outside the sum,
\begin{equation}
    \nabla_{\phi} \mathcal{L}_{\mathrm{DCA}}
    \;=\; -\frac{1}{M} \sum_{k=1}^{K} \frac{\partial M}{\partial \alpha_k}\, \nabla_{\phi} \alpha_k
    \;=\; -\frac{1}{M}\, \nabla_{\phi} M
    \;=\; \frac{1}{M}\, \nabla_{\phi} \mathcal{L}_{\mathrm{TV}}.
    \label{eq:dca-identity}
\end{equation}
The factor is positive at every parameter value and carries no position index, so on this cycle the two objectives share their descent direction, their stationary points, their minimizers and the relative weights they place on the draft positions, the depth weighting of Section~\ref{sec:method-valid} among them.
The logarithm only rescales that cycle's gradient magnitude by the factor $1/M$, common to all positions.

\clearpage
\noindent\textbf{Optimization on a Draft Step.} How much of the factor $M^{-1}$ reaches the parameters depends on the optimizer.
Under SGD with learning rate $\eta$, Eq.~\ref{eq:dca-identity} gives a draft step with chain value $M$ the update $-\eta\, \nabla_{\phi} \mathcal{L}_{\mathrm{DCA}}$ under $\mathcal{L}_{\mathrm{DCA}}$ and the update $-\eta M\, \nabla_{\phi} \mathcal{L}_{\mathrm{DCA}}$ under $\mathcal{L}_{\mathrm{TV}}$.
The rate at which the shared direction acts on the parameters is therefore $\eta$ under $\mathcal{L}_{\mathrm{DCA}}$ and $\eta M \in (0, \eta K]$ under $\mathcal{L}_{\mathrm{TV}}$, which makes the effective rate of $\mathcal{L}_{\mathrm{TV}}$ a function of the head's own acceptance on that draft step and leaves the effective rate of $\mathcal{L}_{\mathrm{DCA}}$ at the scheduled learning rate, without extra scaling by the value $M$.

Adam divides the first moment of the gradient history by the square root of the second moment, and a factor common to every gradient in that history scales the first moment and the square root of the second moment alike, so the factor cancels in the quotient up to the constant that stabilizes the denominator.
Ignoring Adam's stabilizing constant, a positive constant scaling of the entire gradient history cancels in the moment ratio; factors varying across draft cycles generally do not cancel.

\noindent\textbf{Optimization on a Batch.} The optimizer does not receive the gradient of a single draft step: what reaches it is the batch average, in which $M^{-1}$ is no longer a scalar but a weight carried by each draft step separately.
Writing $g_r \triangleq \nabla_{\phi}[-\log M_r]$ for the gradient $\mathcal{L}_{\mathrm{DCA}}$ produces on draft step $r$, Eq.~\ref{eq:dca-identity} makes the gradient of $\mathcal{L}_{\mathrm{TV}}$ on the same draft step $M_r g_r$, giving the following two batch gradients:
\begin{equation}
    \nabla_{\phi}\, \mathbb{E}_r\bigl[\mathcal{L}_{\mathrm{TV}}\bigr] \;=\; \mathbb{E}_r\bigl[M_r\, g_r\bigr],
    \qquad
    \nabla_{\phi}\, \mathbb{E}_r\bigl[\mathcal{L}_{\mathrm{DCA}}\bigr] \;=\; \mathbb{E}_r\bigl[g_r\bigr],
    \label{eq:dca-batch}
\end{equation}
where $\mathbb{E}_r$ averages over the recorded draft steps that make up the batch.
Relative to the DCA cycle gradients $g_r$, the TV-chain objective assigns the explicit weights $M_r$; DCA averages $g_r$ without these factors.
The norms of $g_r$ can still differ across cycles.
Consequently, the stationary conditions $\mathbb{E}_r[M_r g_r]=0$ and $\mathbb{E}_r[g_r]=0$ need not coincide.
The logarithm thus changes the batch objective, favoring smaller-$M_r$ cycles in relative gradient weight without guaranteeing faster convergence.

\noindent\textbf{Conclusion.} The logarithm preserves the single-cycle gradient direction and relative depth weights while changing the batch objective through cycle reweighting.
The experiments assess whether this change in relative cycle weights improves realized acceptance and reduces overall RL training time.

\section{Limitations}
\label{app:limitations}
\noindent\textbf{Run length and compute allocation.}
Net savings depend on the run length and the rollout share of total compute.
In both from-scratch experiments, per-step acceleration begins within 30 steps, and the 500-step runs recover the initial cost with substantial net savings.
Typical RL training spans hundreds or thousands of steps, providing ample time to amortize this brief initial training overhead.

\noindent\textbf{Distribution specialization.}
The grown head is tailored to the rollout distribution of the current RL run.
As shown in \textbf{Appendix~\ref{app:disc-serving}}, it retains higher acceptance within its training domain than across domains.
Its reuse across unrelated tasks therefore requires separate evaluation, while broad deployment remains a distinct setting from the acceleration studied within a single ongoing RL run.

\noindent\textbf{Evaluation scope.}
Our measurements cover 4B--7B models on mathematical and code reasoning with a single node of eight GPUs.
Larger models, multi-node systems, and asynchronous RL may have different compute and communication costs, so their speedups require direct measurement.
The projections in \textbf{Appendix~\ref{app:disc-scale}} illustrate scaling under assumptions, rather than measured performance.

%% file: tabs/notations.tex
\begin{center}
\renewcommand{\arraystretch}{1.15}
\begin{longtable}{p{0.25\linewidth} p{0.68\linewidth}}
\toprule
\textbf{Symbol} & \textbf{Definition} \\
\midrule
\endfirsthead

\toprule
\textbf{Symbol} & \textbf{Definition} \\
\midrule
\endhead

\endfoot

\bottomrule
\\[-0.2em]
\caption{Notation used throughout the paper.}
\label{tab:notation} \\
\endlastfoot

\multicolumn{2}{c}{\textbf{RL Training Notation}} \\
\midrule
\(\pi_\theta\) & The policy, the language model itself. \\
\(\theta\) & Backbone (policy) parameters. \\
\(x\) & Prompt. \\
\(y,\ y_t\) & A sampled response and its token at position \(t\). \\
\(G\) & Number of responses sampled per prompt. \\
\(t\) & RL step index in \autoref{app:lag}; elsewhere the token position of \(y_t\). \\
\(p^{(t)},\ p^{(t+1)}\) & Target distributions before and after one policy update. \\
\(\rho(y)\) & Importance ratio \(p^{(t+1)}(y\mid s)\,/\,p^{(t)}(y\mid s)\) between successive policies. \\
\(\bar{\varepsilon}_t\) & Uniform bound on the per-update target shift. \\

\midrule
\multicolumn{2}{c}{\textbf{Speculative Decoding Notation}} \\
\midrule
\(K\) & Drafting depth: the number of draft tokens per cycle. \\
\(r\) & Draft-then-verify cycle index. \\
\(d_{r,k}\) & The \(k\)-th draft token of cycle \(r\). \\
\(\tilde{d}_{r,j}\) & The replacement token resampled at the first rejection. \\
\(j,\ j_r\) & First-rejection position of a cycle, with \(j=K{+}1\) when every draft is accepted. \\
\(p_{r,k}\) & Target distribution at draft position \(k\) of cycle \(r\). \\
\(q_{r,k}\) & Head distribution at draft position \(k\) of cycle \(r\). \\
\(\alpha_k\) & Acceptance probability at position \(k\): \(\alpha_k = 1 - \mathrm{TV}(p_k, q_k)\). \\
\(\tilde{\alpha}(s),\ \alpha(s)\) & Acceptance rate against the old and the new target at state \(s\). \\
\(\tau\) & Acceptance length: the average number of tokens committed per cycle, \(\tau \in [1, K{+}1]\). \\
\(\mathrm{TV}\) & Total variation distance. \\

\midrule
\multicolumn{2}{c}{\textbf{Model and State Notation}} \\
\midrule
\(z_\theta\) & Target-model logit function. \\
\(h_\phi\) & The MTP draft head, mapping a conditioning state to a distribution. \\
\(\phi\) & Draft-head parameters. \\
\(\mathbf{c}_r\) & Committed context of cycle \(r\): the prompt and all tokens committed before the cycle. \\
\(\mathbf{d}_r\) & Draft path of cycle \(r\): \([d_{r,1},\ldots,d_{r,K}]\). \\
\(s_{r,k}\) & Reconstructed state \([\mathbf{c}_r, d_{r,1},\ldots,d_{r,k-1}]\). \\

\midrule
\multicolumn{2}{c}{\textbf{Training Objective Notation}} \\
\midrule
\(\mathcal{L}_{\mathrm{TV}}\) & TV-chain loss: \(-\sum_{l=1}^{K}\prod_{i=1}^{l}\alpha_i\), a differentiable surrogate for \(1-\tau\). \\
\(\mathcal{L}_{\mathrm{DCA}}\) & DCA loss: \(-\log(-\mathcal{L}_{\mathrm{TV}})\). \\
\(\mathcal{L}_{\mathrm{DCA}}^{\mathrm{VGM}}\) & Verify-gated DCA loss: the chain sum truncated at \(\min(j,K)\). \\
\(M\) & The chain quantity \(-\mathcal{L}_{\mathrm{TV}}\) in \autoref{app:dca}. \\
\(D_k\) & Tail term of the gradient weight: \(\sum_{l=k}^{K}\prod_{i=k+1}^{l}\alpha_i \in [1, K{-}k{+}1]\). \\
\(g_r\) & Gradient of \(\mathcal{L}_{\mathrm{DCA}}\) on cycle \(r\) in \autoref{app:dca}. \\

\end{longtable}
\renewcommand{\arraystretch}{1.0}
\end{center}

%% file: tabs/rq1_eval.tex
\begin{table}[h]
    \centering
    \footnotesize
    \setlength{\tabcolsep}{3pt}
    \caption{\textbf{Inference evaluation on Qwen3-4B (500 RL steps, $K{=}5$).} The trained draft head reaches $\tau$ above $2.1$ on every benchmark in both domains, while policy quality (Mean@16 for math, Pass@4 for code) remains on par with the $K{=}0$ baseline, autoregressive decoding without any draft head.}
    \label{tab:rq1-eval}
    \resizebox{\linewidth}{!}{%
    \begin{tabular}{@{}llcccccccccccc@{}}
        \toprule
        & & \multicolumn{6}{c}{\textbf{Math Reasoning Eval}} & \multicolumn{6}{c}{\textbf{Code Reasoning Eval}} \\
        \cmidrule(lr){3-8} \cmidrule(lr){9-14}
        & & \multicolumn{2}{c}{AMC23} & \multicolumn{2}{c}{AIME24} & \multicolumn{2}{c}{AIME25} & \multicolumn{2}{c}{AtCoder} & \multicolumn{2}{c}{LeetCode} & \multicolumn{2}{c}{LiveCodeBench} \\
        \cmidrule(lr){3-4} \cmidrule(lr){5-6} \cmidrule(lr){7-8} \cmidrule(lr){9-10} \cmidrule(lr){11-12} \cmidrule(lr){13-14}
        Method & $K$ & Mean@16 & $\tau$ & Mean@16 & $\tau$ & Mean@16 & $\tau$ & Pass@4 & $\tau$ & Pass@4 & $\tau$ & Pass@4 & $\tau$ \\
        \midrule
        AR RL       & $0$ & $62.81$ & $1.00$ & $20.42$ & $1.00$ & $18.75$ & $1.00$ & $42.19$ & $1.00$ & $42.79$ & $1.00$ & $42.37$ & $1.00$ \\
        \rowcolor{brandgreen}
        \growmtp   & $5$ & $64.69$ & $\mathbf{2.33}$ & $20.42$ & $\mathbf{2.34}$ & $21.04$ & $\mathbf{2.30}$ & $43.36$ & $\mathbf{2.27}$ & $43.24$ & $\mathbf{2.11}$ & $43.51$ & $\mathbf{2.22}$ \\
        \bottomrule
    \end{tabular}%
    }
\end{table}

%% file: tabs/rq2a_eval.tex
\begin{table}[h]
    \centering
    \footnotesize
    \setlength{\tabcolsep}{3pt}
    \caption{\textbf{Inference evaluation on Qwen3.5-4B-Base (200 RL steps, $K{=}3$).} Among quality-preserving strategies, DCA reaches the highest $\tau$ on every benchmark. Joint CE collapses to zero.}
    \label{tab:rq2a-eval}
    \resizebox{\linewidth}{!}{%
    \begin{tabular}{@{}llcccccccccccc@{}}
        \toprule
        & & \multicolumn{6}{c}{\textbf{Math Reasoning Eval}} & \multicolumn{6}{c}{\textbf{Code Reasoning Eval}} \\
        \cmidrule(lr){3-8} \cmidrule(lr){9-14}
        & & \multicolumn{2}{c}{AMC23} & \multicolumn{2}{c}{AIME24} & \multicolumn{2}{c}{AIME25} & \multicolumn{2}{c}{AtCoder} & \multicolumn{2}{c}{LeetCode} & \multicolumn{2}{c}{LiveCodeBench} \\
        \cmidrule(lr){3-4} \cmidrule(lr){5-6} \cmidrule(lr){7-8} \cmidrule(lr){9-10} \cmidrule(lr){11-12} \cmidrule(lr){13-14}
        Training Mode & Loss & Mean@16 & $\tau$ & Mean@16 & $\tau$ & Mean@16 & $\tau$ & Pass@4 & $\tau$ & Pass@4 & $\tau$ & Pass@4 & $\tau$ \\
        \midrule
        Frozen   & --  & $80.47$ & $2.61$ & $55.00$ & $2.57$ & $46.67$ & $2.57$ & $63.29$ & $2.52$ & $72.52$ & $2.52$ & $67.39$ & $2.52$ \\
        Joint    & CE  & $0.00$  & $3.99$ & $0.00$  & $3.99$ & $0.00$  & $3.99$ & $0.00$ & $3.98$ & $0.00$ & $3.98$ & $0.00$ & $3.98$ \\
        Detached & CE  & $84.69$ & $3.19$ & $57.92$ & $3.10$ & $52.50$ & $3.10$ & $61.63$ & $3.15$ & $71.40$ & $3.13$ & $65.97$ & $3.14$ \\
        \rowcolor{brandgreen}
        Detached & DCA (\growmtp) & $85.78$ & $\mathbf{3.31}$ & $59.58$ & $\mathbf{3.23}$ & $51.46$ & $\mathbf{3.22}$ & $63.29$ & $\mathbf{3.22}$ & $70.50$ & $\mathbf{3.22}$ & $66.54$ & $\mathbf{3.22}$ \\
        \bottomrule
    \end{tabular}%
    }
\end{table}

%% file: tabs/rq2b_eval.tex
\begin{table}[h]
    \centering
    \footnotesize
    \setlength{\tabcolsep}{3pt}
    \caption{\textbf{Inference evaluation across drafting depths on MiMo-7B-SFT (200 RL steps).} Online training lifts $\tau$ above $3.0$ at every depth, monotonically in $K$, while policy quality stays on par with the frozen head. The highlighted row follows end-to-end acceleration, not the highest measured $\tau$.}
    \label{tab:rq2b-eval}
    \resizebox{\linewidth}{!}{%
    \begin{tabular}{@{}llcccccccccccc@{}}
        \toprule
        & & \multicolumn{6}{c}{\textbf{Math Reasoning Eval}} & \multicolumn{6}{c}{\textbf{Code Reasoning Eval}} \\
        \cmidrule(lr){3-8} \cmidrule(lr){9-14}
        & & \multicolumn{2}{c}{AMC23} & \multicolumn{2}{c}{AIME24} & \multicolumn{2}{c}{AIME25} & \multicolumn{2}{c}{AtCoder} & \multicolumn{2}{c}{LeetCode} & \multicolumn{2}{c}{LiveCodeBench} \\
        \cmidrule(lr){3-4} \cmidrule(lr){5-6} \cmidrule(lr){7-8} \cmidrule(lr){9-10} \cmidrule(lr){11-12} \cmidrule(lr){13-14}
        Method & $K$ & Mean@16 & $\tau$ & Mean@16 & $\tau$ & Mean@16 & $\tau$ & Pass@4 & $\tau$ & Pass@4 & $\tau$ & Pass@4 & $\tau$ \\
        \midrule
        Frozen    & $1$ & $90.47$ & $1.67$ & $51.88$ & $1.64$ & $43.13$ & $1.64$ & $69.93$ & $1.61$ & $80.18$ & $1.61$ & $74.50$ & $1.61$ \\
        \growmtp & $3$ & $90.47$ & $3.20$ & $53.33$ & $3.15$ & $42.29$ & $3.14$ & $70.10$ & $3.02$ & $78.38$ & $3.03$ & $73.84$ & $3.03$ \\
        \rowcolor{brandgreen}
        \growmtp & $5$ & $91.25$ & $3.95$ & $56.87$ & $3.84$ & $46.25$ & $3.82$ & $71.59$ & $3.59$ & $78.60$ & $3.59$ & $74.69$ & $3.59$ \\
        \growmtp & $7$ & $89.69$ & $4.30$ & $56.04$ & $4.11$ & $43.33$ & $4.08$ & $69.93$ & $3.83$ & $76.35$ & $3.85$ & $72.80$ & $3.84$ \\
        \bottomrule
    \end{tabular}%
    }
\end{table}

%% file: tabs/rq3_dca.tex
\begin{table}[h]
    \centering
    \footnotesize
    \setlength{\tabcolsep}{3pt}
    \caption{\textbf{Ablation of Objective on MiMo-7B-SFT (200 RL steps, $K\in\{1,5\}$, mathematical reasoning).} At $K{=}1$, the four objectives yield similar acceptance lengths. At $K{=}5$, DCA leads on training acceptance, end-to-end step speedup, and held-out $\tau$ on each of the three benchmarks.}
    \label{tab:rq3-dca}
    \resizebox{\linewidth}{!}{%
    \begin{tabular}{@{}llccccccccccc@{}}
        \toprule
        & & \multicolumn{5}{c}{\textbf{Math Reasoning RL}} & \multicolumn{6}{c}{\textbf{Math Reasoning Eval}} \\
        \cmidrule(lr){3-7} \cmidrule(lr){8-13}
        & & \multicolumn{5}{c}{DAPO-Math-17K} & \multicolumn{2}{c}{AMC23} & \multicolumn{2}{c}{AIME24} & \multicolumn{2}{c}{AIME25} \\
        \cmidrule(lr){3-7} \cmidrule(lr){8-9} \cmidrule(lr){10-11} \cmidrule(lr){12-13}
        Loss & $K$ & Rollout (s) & Speedup & Step (s) & Speedup & $\tau$ & Mean@16 & $\tau$ & Mean@16 & $\tau$ & Mean@16 & $\tau$ \\
        \midrule
        CE  & $1$ & $274.47$ & $1.16\times$ & $507.99$ & $1.03\times$ & $1.84$ & $90.16$ & $1.88$ & $51.67$ & $1.87$ & $40.42$ & $1.87$ \\
        KL  & $1$ & $288.58$ & $1.10\times$ & $532.91$ & $0.98\times$ & $1.81$ & $90.31$ & $1.86$ & $54.37$ & $1.85$ & $44.58$ & $1.84$ \\
        TV  & $1$ & $288.30$ & $1.11\times$ & $530.29$ & $0.99\times$ & $1.83$ & $89.22$ & $1.88$ & $56.04$ & $1.86$ & $43.13$ & $1.86$ \\
        \rowcolor{brandgreen}
        DCA (\growmtp) & $1$ & $280.58$ & $1.14\times$ & $517.00$ & $1.01\times$ & $1.82$ & $89.53$ & $1.87$ & $53.96$ & $1.85$ & $36.88$ & $1.86$ \\
        \midrule
        CE  & $5$ & $175.70$ & $1.81\times$ & $384.22$ & $1.36\times$ & $3.86$ & $89.84$ & $3.81$ & $54.17$ & $3.67$ & $43.54$ & $3.66$ \\
        KL  & $5$ & $174.98$ & $1.82\times$ & $382.41$ & $1.37\times$ & $3.85$ & $89.69$ & $3.67$ & $53.75$ & $3.53$ & $42.29$ & $3.51$ \\
        TV  & $5$ & $174.00$ & $1.83\times$ & $382.69$ & $1.37\times$ & $3.94$ & $89.53$ & $3.80$ & $55.00$ & $3.67$ & $40.83$ & $3.66$ \\
        \rowcolor{brandgreen}
        DCA (\growmtp) & $5$ & $\mathbf{164.84}$ & $\mathbf{1.93\times}$ & $\mathbf{371.02}$ & $\mathbf{1.41\times}$ & $\mathbf{4.04}$ & $91.25$ & $\mathbf{3.95}$ & $56.87$ & $\mathbf{3.84}$ & $46.25$ & $\mathbf{3.82}$ \\
        \bottomrule
    \end{tabular}%
    }
\end{table}

%% file: tabs/rq3_vgm.tex
\begin{table}[h]
    \centering
    \footnotesize
    \setlength{\tabcolsep}{3pt}
    \caption{\textbf{Ablation of Verify-Gated Masking on MiMo-7B-SFT (200 RL steps, $K{=}5$, mathematical reasoning).} Removing VGM slows end-to-end training steps under both objectives, and KL degrades more than DCA: the step speedup falls from $1.37\times$ to $1.28\times$ and from $1.41\times$ to $1.35\times$.}
    \label{tab:rq3-vgm}
    \resizebox{\linewidth}{!}{%
    \begin{tabular}{@{}lcccccccccccc@{}}
        \toprule
        & & \multicolumn{5}{c}{\textbf{Math Reasoning RL}} & \multicolumn{6}{c}{\textbf{Math Reasoning Eval}} \\
        \cmidrule(lr){3-7} \cmidrule(lr){8-13}
        & & \multicolumn{5}{c}{DAPO-Math-17K} & \multicolumn{2}{c}{AMC23} & \multicolumn{2}{c}{AIME24} & \multicolumn{2}{c}{AIME25} \\
        \cmidrule(lr){3-7} \cmidrule(lr){8-9} \cmidrule(lr){10-11} \cmidrule(lr){12-13}
        Loss & VGM & Rollout (s) & Speedup & Step (s) & Speedup & $\tau$ & Mean@16 & $\tau$ & Mean@16 & $\tau$ & Mean@16 & $\tau$ \\
        \midrule
        KL & \ding{55} & $186.40$ & $1.71\times$ & $408.65$ & $1.28\times$ & $3.80$ & $88.59$ & $3.66$ & $54.17$ & $3.49$ & $41.04$ & $3.48$ \\
        \rowcolor{brandgreen}
        KL & \ding{51} & $174.98$ & $1.82\times$ & $382.41$ & $1.37\times$ & $3.85$ & $89.69$ & $3.67$ & $53.75$ & $3.53$ & $42.29$ & $3.51$ \\
        \midrule
        DCA & \ding{55} & $165.97$ & $1.92\times$ & $387.27$ & $1.35\times$ & $4.00$ & $89.22$ & $3.94$ & $52.92$ & $3.81$ & $41.88$ & $3.79$ \\
        \rowcolor{brandgreen}
        DCA (\growmtp) & \ding{51} & $164.84$ & $1.93\times$ & $\mathbf{371.02}$ & $\mathbf{1.41\times}$ & $4.04$ & $91.25$ & $3.95$ & $56.87$ & $3.84$ & $46.25$ & $3.82$ \\
        \bottomrule
    \end{tabular}%
    }
\end{table}

%% file: tabs/disc_scale.tex
\begin{wraptable}{r}{0.50\linewidth}
    \centering
    \small
    \setlength{\tabcolsep}{6pt}
    \caption{\textbf{The projected speedup of \growmtp grows monotonically with RL scale.} End-to-end step speedup over the AR baseline, extrapolated from the measured Qwen3-4B run on DAPO-Math-17K ($K{=}5$, the 500-step 8K entry).}
    \label{tab:disc-scale}
    \begin{tabular}{@{}lccc@{}}
        \toprule
        & \multicolumn{3}{c}{Rollout Length} \\
        \cmidrule(lr){2-4}
        Math Reasoning RL Steps & 8K & 16K & 32K \\
        \midrule
        $500$  & $1.60\times$ & $1.79\times$ & $1.93\times$ \\
        $1000$ & $1.70\times$ & $1.91\times$ & $2.08\times$ \\
        $2000$ & $1.76\times$ & $1.99\times$ & $2.17\times$ \\
        \bottomrule
    \end{tabular}
\end{wraptable}

%% file: tabs/disc_cross_domain.tex
\begingroup
    \captionsetup{type=table,hypcap=false,belowskip=5pt}
    \centering
    \small
    \setlength{\tabcolsep}{3pt}
    \caption{\textbf{Each head specializes to the domain it was trained on.} Acceptance length $\tau$ on \mbox{Qwen3-4B} ($K{=}5$), with Rollout on each head's own training rollouts and shaded blocks in-domain.}
    \label{tab:disc-cross-domain}
    \resizebox{\linewidth}{!}{%
    \begin{tabular}{@{}lccccccccc@{}}
        \toprule
        & & \multicolumn{4}{c}{\textbf{Math Reasoning Eval $\tau$}} & \multicolumn{4}{c}{\textbf{Code Reasoning Eval $\tau$}} \\
        \cmidrule(lr){3-6} \cmidrule(lr){7-10}
        Training Domain & Rollout & AMC23 & AIME24 & AIME25 & Avg. & AtCoder & LeetCode & LiveCodeBench & Avg. \\
        \midrule
        Math & $2.91$ & \cellcolor{brandgreen}$2.33$ & \cellcolor{brandgreen}$2.34$ & \cellcolor{brandgreen}$2.30$ & \cellcolor{brandgreen}$2.32$ & $1.50$ & $1.40$ & $1.46$ & $1.45$ \\
        Code & $2.64$ & $1.80$ & $1.79$ & $1.68$ & $1.75$ & \cellcolor{brandgreen}$2.27$ & \cellcolor{brandgreen}$2.11$ & \cellcolor{brandgreen}$2.22$ & \cellcolor{brandgreen}$2.20$ \\
        \bottomrule
    \end{tabular}%
    }
    \par
\endgroup

%% file: style/iclr2027_conference.bib
@article{jaech2024openai,
  title={Openai o1 system card},
  author={Jaech, Aaron and Kalai, Adam and Lerer, Adam and Richardson, Adam and El-Kishky, Ahmed and Low, Aiden and Helyar, Alec and Madry, Aleksander and Beutel, Alex and Carney, Alex and others},
  journal={arXiv preprint arXiv:2412.16720},
  year={2024}
}

@article{guo2025deepseek,
  title={Deepseek-r1: Incentivizing reasoning capability in llms via reinforcement learning},
  author={Guo, Daya and Yang, Dejian and Zhang, Haowei and Song, Junxiao and Wang, Peiyi and Zhu, Qihao and Xu, Runxin and Zhang, Ruoyu and Ma, Shirong and Bi, Xiao and others},
  journal={arXiv preprint arXiv:2501.12948},
  year={2025}
}

@article{team2025kimi,
  title={Kimi k1. 5: Scaling reinforcement learning with llms, 2025},
  author={Team, Kimi and Du, Angang and Gao, Bofei and Xing, Bowei and Jiang, Changjiu and Chen, Cheng and Li, Cheng and Xiao, Chenjun and Du, C and Liao, C and others},
  journal={URL https://arxiv. org/abs/2501.12599},
  volume={118},
  year={2025}
}

@article{singh2025openai,
  title={Openai gpt-5 system card},
  author={Singh, Aaditya and Fry, Adam and Perelman, Adam and Tart, Adam and Ganesh, Adi and El-Kishky, Ahmed and McLaughlin, Aidan and Low, Aiden and Ostrow, AJ and Ananthram, Akhila and others},
  journal={arXiv preprint arXiv:2601.03267},
  year={2025}
}

@inproceedings{khatri2026art,
  title={The art of scaling reinforcement learning compute for llms},
  author={Khatri, Devvrit and Madaan, Lovish and Tiwari, Rishabh and Bansal, Rachit and Duvvuri, Venkata Sai Surya Subramanyam and Zaheer, Manzil and Dhillon, Inderjit and Brandfonbrener, David and Agarwal, Rishabh},
  booktitle={International Conference on Learning Representations},
  volume={2026},
  pages={72438--72467},
  year={2026}
}

@article{chen2026respec,
  title={Respec: Towards optimizing speculative decoding in reinforcement learning systems},
  author={Chen, Qiaoling and Liu, Zijun and Sun, Peng and Li, Shenggui and Wang, Guoteng and Liu, Ziming and Wen, Yonggang and Feng, Siyuan and Zhang, Tianwei},
  journal={Proceedings of Machine Learning and Systems},
  volume={8},
  pages={367--379},
  year={2026}
}

@inproceedings{leviathan2023fast,
  title={Fast inference from transformers via speculative decoding},
  author={Leviathan, Yaniv and Kalman, Matan and Matias, Yossi},
  booktitle={International conference on machine learning},
  pages={19274--19286},
  year={2023},
  organization={PMLR}
}

@article{chen2023accelerating,
  title={Accelerating large language model decoding with speculative sampling},
  author={Chen, Charlie and Borgeaud, Sebastian and Irving, Geoffrey and Lespiau, Jean-Baptiste and Sifre, Laurent and Jumper, John},
  journal={arXiv preprint arXiv:2302.01318},
  year={2023}
}

@article{cai2024medusa,
  title={Medusa: Simple llm inference acceleration framework with multiple decoding heads},
  author={Cai, Tianle and Li, Yuhong and Geng, Zhengyang and Peng, Hongwu and Lee, Jason D and Chen, Deming and Dao, Tri},
  journal={arXiv preprint arXiv:2401.10774},
  year={2024}
}

@article{li2024eagle,
  title={Eagle: Speculative sampling requires rethinking feature uncertainty},
  author={Li, Yuhui and Wei, Fangyun and Zhang, Chao and Zhang, Hongyang},
  journal={arXiv preprint arXiv:2401.15077},
  year={2024}
}

@article{li2026eagle,
  title={Eagle-3: Scaling up inference acceleration of large language models via training-time test},
  author={Li, Yuhui and Wei, Fangyun and Zhang, Chao and Zhang, Hongyang},
  journal={Advances in Neural Information Processing Systems},
  volume={38},
  pages={136737--136756},
  year={2026}
}

@article{liu2024deepseek,
  title={Deepseek-v3 technical report},
  author={Liu, Aixin and Feng, Bei and Xue, Bing and Wang, Bingxuan and Wu, Bochao and Lu, Chengda and Zhao, Chenggang and Deng, Chengqi and Zhang, Chenyu and Ruan, Chong and others},
  journal={arXiv preprint arXiv:2412.19437},
  year={2024}
}

@article{xiaomi2025mimo,
  title={MiMo: Unlocking the Reasoning Potential of Language Model--From Pretraining to Posttraining},
  author={Xiaomi, LLM and Xia, Bingquan and Shen, Bowen and Zhu, Dawei and Zhang, Di and Wang, Gang and Zhang, Hailin and Liu, Huaqiu and Xiao, Jiebao and Dong, Jinhao and others},
  journal={arXiv preprint arXiv:2505.07608},
  year={2025}
}

@article{gloeckle2024better,
  title={Better \& faster large language models via multi-token prediction},
  author={Gloeckle, Fabian and Idrissi, Badr Youbi and Rozi{\`e}re, Baptiste and Lopez-Paz, David and Synnaeve, Gabriel},
  journal={arXiv preprint arXiv:2404.19737},
  year={2024}
}

@inproceedings{hu2026bridging,
  title={Bridging draft policy misalignment: Group tree optimization for speculative decoding},
  author={Hu, Shijing and Li, Jingyang and Lu, Zhihui and Zhou, Pan},
  booktitle={International Conference on Learning Representations},
  volume={2026},
  pages={112507--112524},
  year={2026}
}

@article{ankner2024hydra,
  title={Hydra: Sequentially-dependent draft heads for medusa decoding},
  author={Ankner, Zachary and Parthasarathy, Rishab and Nrusimha, Aniruddha and Rinard, Christopher and Ragan-Kelley, Jonathan and Brandon, William},
  journal={arXiv preprint arXiv:2402.05109},
  year={2024}
}

@inproceedings{li2024eagle2,
  title={Eagle-2: Faster inference of language models with dynamic draft trees},
  author={Li, Yuhui and Wei, Fangyun and Zhang, Chao and Zhang, Hongyang},
  booktitle={Proceedings of the 2024 conference on empirical methods in natural language processing},
  pages={7421--7432},
  year={2024}
}

@article{cao2026qwen3,
  title={Qwen3-coder-next technical report},
  author={Cao, Ruisheng and Chen, Mouxiang and Chen, Jiawei and Cui, Zeyu and Feng, Yunlong and Hui, Binyuan and Jing, Yuheng and Li, Kaixin and Li, Mingze and Lin, Junyang and others},
  journal={arXiv preprint arXiv:2603.00729},
  year={2026}
}

@article{yang2025qwen3,
  title={Qwen3 technical report},
  author={Yang, An and Li, Anfeng and Yang, Baosong and Zhang, Beichen and Hui, Binyuan and Zheng, Bo and Yu, Bowen and Gao, Chang and Huang, Chengen and Lv, Chenxu and others},
  journal={arXiv preprint arXiv:2505.09388},
  year={2025}
}

@misc{qwen35blog,
    title = {Qwen3.5: Towards Native Multimodal Agents},
    url = {https://qwen.ai/blog?id=qwen3.5},
    author = {{Qwen}},
    month = {February},
    year = {2026}
}

@article{xia2024unlocking,
  title={Unlocking efficiency in large language model inference: A comprehensive survey of speculative decoding},
  author={Xia, Heming and Yang, Zhe and Dong, Qingxiu and Wang, Peiyi and Li, Yongqi and Ge, Tao and Liu, Tianyu and Li, Wenjie and Sui, Zhifang},
  journal={Findings of the Association for Computational Linguistics: ACL 2024},
  pages={7655--7671},
  year={2024}
}

@inproceedings{miao2024specinfer,
  title={Specinfer: Accelerating large language model serving with tree-based speculative inference and verification},
  author={Miao, Xupeng and Oliaro, Gabriele and Zhang, Zhihao and Cheng, Xinhao and Wang, Zeyu and Zhang, Zhengxin and Wong, Rae Ying Yee and Zhu, Alan and Yang, Lijie and Shi, Xiaoxiang and others},
  booktitle={Proceedings of the 29th ACM International Conference on Architectural Support for Programming Languages and Operating Systems, Volume 3},
  pages={932--949},
  year={2024}
}

@article{gspo,
  title={Group Sequence Policy Optimization}, 
  author={
    Chujie Zheng and Shixuan Liu and Mingze Li and Xiong-Hui Chen and Bowen Yu and 
    Chang Gao and Kai Dang and Yuqiong Liu and Rui Men and An Yang and Jingren Zhou and 
    Junyang Lin 
  },
  journal={arXiv preprint arXiv:2507.18071},
  year={2025}
}

@article{xu2026deepseek,
  title={Deepseek-v4: Towards highly efficient million-token context intelligence},
  author={Xu, Anyi and Lin, Bangcai and Xue, Bing and Wang, Bingxuan and Xu, Bingzheng and Wu, Bochao and Zhang, Bowei and Lin, Chaofan and Dong, Chen and Ling, Chenchen and others},
  journal={arXiv preprint arXiv:2606.19348},
  year={2026}
}

@article{li2026breaking,
  title={Breaking Entropy Bounds: Accelerating RL Training via MTP with Rejection Sampling},
  author={Li, Yucheng and Jiang, Huiqiang and Xu, Yang and Yang, Jianxin and Zhang, Yi and Cao, Yizhong and Shen, Yuhao and Zhou, Fan and Men, Rui and Zhang, Jianwei and others},
  journal={arXiv preprint arXiv:2606.12370},
  year={2026}
}

@article{yu2026dapo,
  title={Dapo: An open-source llm reinforcement learning system at scale},
  author={Yu, Qiying and Zhang, Zheng and Zhu, Ruofei and Yuan, Yufeng and Zuo, Xiaochen and Yue, Yu and Dai, Weinan and Fan, Tiantian and Liu, Gaohong and Liu, Lingjun and others},
  journal={Advances in Neural Information Processing Systems},
  volume={38},
  pages={113222--113244},
  year={2026}
}

@article{li2023taco,
  title={Taco: Topics in algorithmic code generation dataset},
  author={Li, Rongao and Fu, Jie and Zhang, Bo-Wen and Huang, Tao and Sun, Zhihong and Lyu, Chen and Liu, Guang and Jin, Zhi and Li, Ge},
  journal={arXiv preprint arXiv:2312.14852},
  year={2023}
}

@inproceedings{jain2025livecodebench,
  title={Livecodebench: Holistic and contamination free evaluation of large language models for code},
  author={Jain, Naman and Gu, Alex and Li, Wen-Ding and Yan, Fanjia and Zhang, Tianjun and Wang, Sida and Solar-Lezama, Armando and Sen, Koushik and Stoica, Ion},
  booktitle={International Conference on Learning Representations},
  volume={2025},
  pages={58791--58831},
  year={2025}
}

@misc{aime24,
      title={American Invitational Mathematics Examination (AIME) 2024}, 
      author={Zhang, Yifan and Math-AI, Team},
      year={2024},
}

@misc{aime25,
      title={American Invitational Mathematics Examination (AIME) 2025}, 
      author={Zhang, Yifan and Math-AI, Team},
      year={2025},
}

@misc{amc23,
      title={American Mathematics Competitions (AMC) 2023},
      author={Zhang, Yifan and Math-AI, Team},
      year={2023},
}

@article{li2026scaling,
  title={Scaling Data Difficulty: Improving Coding Models via Reinforcement Learning on Fresh and Challenging Problems},
  author={Li, Zongqian and Lv, Tengchao and Huang, Shaohan and Su, Yixuan and Sun, Qinzheng and Yin, Qiufeng and Xin, Ying and Li, Scarlett and Cui, Lei and Collier, Nigel and others},
  journal={arXiv preprint arXiv:2603.07779},
  year={2026}
}

@inproceedings{sheng2025hybridflow,
  title={Hybridflow: A flexible and efficient rlhf framework},
  author={Sheng, Guangming and Zhang, Chi and Ye, Zilingfeng and Wu, Xibin and Zhang, Wang and Zhang, Ru and Peng, Yanghua and Lin, Haibin and Wu, Chuan},
  booktitle={Proceedings of the Twentieth European Conference on Computer Systems},
  pages={1279--1297},
  year={2025}
}

@article{zheng2024sglang,
  title={Sglang: Efficient execution of structured language model programs},
  author={Zheng, Lianmin and Yin, Liangsheng and Xie, Zhiqiang and Sun, Chuyue and Huang, Jeff and Yu, Cody H and Cao, Shiyi and Kozyrakis, Christos and Stoica, Ion and Gonzalez, Joseph E and others},
  journal={Advances in neural information processing systems},
  volume={37},
  pages={62557--62583},
  year={2024}
}

@article{wang2026joint,
  title={Joint Training of Multi-Token Prediction in Reinforcement Learning via Optimal Coefficient Calibration},
  author={Wang, Zili and Chai, Jiajun and Chen, Lin and Wang, Xiaohan and Xiang, Shiming and Yin, Guojun},
  journal={arXiv preprint arXiv:2605.28184},
  year={2026}
}

@article{chandiramani2026nemotron,
  title={Nemotron 3 super: Open, efficient mixture-of-experts hybrid mamba-transformer model for agentic reasoning},
  author={Chandiramani, Aakshita and Blakeman, Aaron and Olaoye, Abdullahi and Gupta, Abhibha and Somasamudramath, Abhilash and Khattar, Abhinav and Adesoba, Adeola and Renduchintala, Adi and Asif, Adil and Agrawal, Aditya and others},
  journal={arXiv preprint arXiv:2604.12374},
  year={2026}
}

@article{zeng2025glm,
  title={Glm-4.5: Agentic, reasoning, and coding (arc) foundation models},
  author={Zeng, Aohan and Lv, Xin and Zheng, Qinkai and Hou, Zhenyu and Chen, Bin and Xie, Chengxing and Wang, Cunxiang and Yin, Da and Zeng, Hao and Zhang, Jiajie and others},
  journal={arXiv preprint arXiv:2508.06471},
  year={2025}
}

@article{chen2024sequoia,
  title={Sequoia: Scalable, robust, and hardware-aware speculative decoding},
  author={Chen, Zhuoming and May, Avner and Svirschevski, Ruslan and Huang, Yuhsun and Ryabinin, Max and Jia, Zhihao and Chen, Beidi},
  journal={arXiv preprint arXiv:2402.12374},
  year={2024}
}

@inproceedings{zhang2025learning,
  title={Learning harmonized representations for speculative sampling},
  author={Zhang, Lefan and Wang, Xiaodan and Huang, Yanhua and Xu, Ruiwen},
  booktitle={International Conference on Learning Representations},
  volume={2025},
  pages={35367--35388},
  year={2025}
}

@article{liu2023online,
  title={Online speculative decoding},
  author={Liu, Xiaoxuan and Hu, Lanxiang and Bailis, Peter and Cheung, Alvin and Deng, Zhijie and Stoica, Ion and Zhang, Hao},
  journal={arXiv preprint arXiv:2310.07177},
  year={2023}
}

@article{lei2026draft,
  title={Draft-OPD: On-Policy Distillation for Speculative Draft Models},
  author={Lei, Haodi and Li, Yafu and Zhang, Haoran and Zhang, Shunkai and Cheng, Qianjia and Qu, Xiaoye and Cui, Ganqu and Zhou, Bowen and Ding, Ning and Luo, Yun and others},
  journal={arXiv preprint arXiv:2605.29343},
  year={2026}
}

@article{fu2026areal,
  title={Areal: A large-scale asynchronous reinforcement learning system for language reasoning},
  author={Fu, Wei and Gao, Jiaxuan and Shen, Xujie and Zhu, Chen and Mei, Zhiyu and He, Chuyi and Xu, Shusheng and Wei, Guo and Mei, Jun and Wang, Jiashu and others},
  journal={Advances in Neural Information Processing Systems},
  volume={38},
  pages={36256--36282},
  year={2026}
}

@article{he2025history,
  title={History rhymes: Accelerating llm reinforcement learning with rhymerl},
  author={He, Jingkai and Li, Tianjian and Feng, Erhu and Du, Dong and Liu, Qian and Liu, Tao and Xia, Yubin and Chen, Haibo},
  journal={arXiv preprint arXiv:2508.18588},
  year={2025}
}

@inproceedings{hu2025openrlhf,
  title={OpenRLHF: A ray-based easy-to-use, scalable and high-performance rlhf framework},
  author={Hu, Jian and Wu, Xibin and Shen, Wei and Liu, Jason Klein and Wang, Weixun and Jiang, Songlin and Wang, Haoran and Chen, Hao and Chen, Bin and Fang, Wenkai and others},
  booktitle={Proceedings of the 2025 Conference on Empirical Methods in Natural Language Processing: System Demonstrations},
  pages={656--666},
  year={2025}
}

@article{shen2024nemo,
  title={Nemo-aligner: Scalable toolkit for efficient model alignment},
  author={Shen, Gerald and Wang, Zhilin and Delalleau, Olivier and Zeng, Jiaqi and Dong, Yi and Egert, Daniel and Sun, Shengyang and Zhang, Jimmy and Jain, Sahil and Taghibakhshi, Ali and others},
  journal={arXiv preprint arXiv:2405.01481},
  year={2024}
}

@inproceedings{noukhovitch2025asynchronous,
  title={Asynchronous rlhf: Faster and more efficient off-policy rl for language models},
  author={Noukhovitch, Michael and Huang, Shengyi and Xhonneux, Sophie and Hosseini, Arian and Agarwal, Rishabh and Courville, Aaron},
  booktitle={International Conference on Learning Representations},
  volume={2025},
  pages={4003--4029},
  year={2025}
}

@inproceedings{zhou2024distillspec,
  title={Distillspec: Improving speculative decoding via knowledge distillation},
  author={Zhou, Yongchao and Lyu, Kaifeng and Rawat, Ankit Singh and Menon, Aditya Krishna and Rostamizadeh, Afshin and Kumar, Sanjiv and Kagy, Jean-Fran{\c{c}}ois and Agarwal, Rishabh},
  booktitle={International Conference on Learning Representations},
  volume={2024},
  pages={32011--32050},
  year={2024}
}
